\documentclass[10pt,twocolumn,letterpaper]{article}

\usepackage{iccv}
\usepackage{times}
\usepackage{url}
\usepackage{epsfig}
\usepackage{graphicx}
\usepackage{amsmath}
\usepackage{amssymb}
\usepackage{multirow}
\usepackage{float}
\usepackage{enumitem}
\usepackage{pifont}
\usepackage{booktabs}
\usepackage{caption}
\usepackage{subcaption}
\usepackage{colortbl}
\usepackage[dvipsnames]{xcolor}

\def\ourmodel{MV-Map\xspace}
\newcommand{\cmark}{\ding{51}}%

\newmuskip\normalthickmuskip
\newmuskip\normalthinmuskip
\definecolor{lightgreen}{HTML}{D8ECD1}
\AtBeginDocument{%
  \normalthickmuskip=\thickmuskip
  \normalthinmuskip=\thinmuskip
}
\everymath{%
  \thickmuskip=\muexpr\normalthickmuskip-(1\normalthinmuskip plus 1\normalthinmuskip)\relax
}
\everydisplay{\thickmuskip=\normalthickmuskip}
\let\texdisplaystyle\displaystyle
\renewcommand{\displaystyle}{\texdisplaystyle\the\everydisplay}
\usepackage[pagebackref=true,breaklinks=true,letterpaper=true,colorlinks,bookmarks=false,hypertexnames=false]{hyperref}

\iccvfinalcopy 
\def\iccvPaperID{} 

\newcommand\mypar[1]{\par\vspace{1.0mm}\noindent\textbf{#1}\;\;}

\def\ourmodel{MV-Map\xspace}
\def\ourmodelfull{Multi-view Map\xspace}

\begin{document}

\title{MV-Map: Offboard HD Map Generation with Multi-view Consistency}

\author{Ziyang Xie$^{*1,2}$ \quad\quad\ \quad\quad Ziqi Pang$^{*2}$\quad\quad Yu-Xiong Wang$^2$\\
Fudan University$^1$ \quad \quad University of Illinois Urbana-Champaign$^2$
\\
{\tt\small ziyangxie19@fudan.edu.cn, ziqip2@illinois.edu, yxw@illinois.edu}
}

\maketitle
\def\thefootnote{*}\footnotetext{Equal contribution.}
\def\thefootnote{\arabic{footnote}}
\ificcvfinal\thispagestyle{empty}\fi

\newcommand{\zy}[1]{\textcolor{orange}{[{\bf ziyang:} #1]}}

\begin{abstract}

While bird's-eye-view (BEV) perception models can be helpful in building high-definition maps (HD Maps) with less human labor, their results are often unreliable and demonstrate noticeable inconsistencies in the predicted HD Maps from different viewpoints. This is because BEV perception is typically set up in an ``onboard'' manner, which restricts the computation and prevents algorithms from simultaneously reasoning multiple views. This paper overcomes these limitations and advocates a more practical ``offboard'' HD Map generation setup that removes the computation constraints, based on the fact that HD Maps are commonly reusable infrastructures built offline in data centers. To this end, we propose a novel offboard pipeline called \ourmodel that capitalizes multi-view consistency and can handle an arbitrary number of frames with the key design of a ``region-centric'' framework. In \ourmodel, the target HD Maps are created by aggregating all the frames of onboard predictions, weighted by the confidence scores assigned by an ``uncertainty network.'' To further enhance multi-view consistency, we augment the uncertainty network with the global 3D structure optimized by a voxelized neural radiance field (Voxel-NeRF). Extensive experiments on nuScenes show that our \ourmodel significantly improves the quality of HD-Maps, further highlighting the importance of offboard methods for HD-Map generation. 
Our code and model are available at \url{https://github.com/ZiYang-xie/MV-Map}.
\end{abstract}

\ificcvfinal\pagestyle{empty}\fi
\section{Introduction}
\begin{figure}
    \centering
    \includegraphics[width=1.0\linewidth]{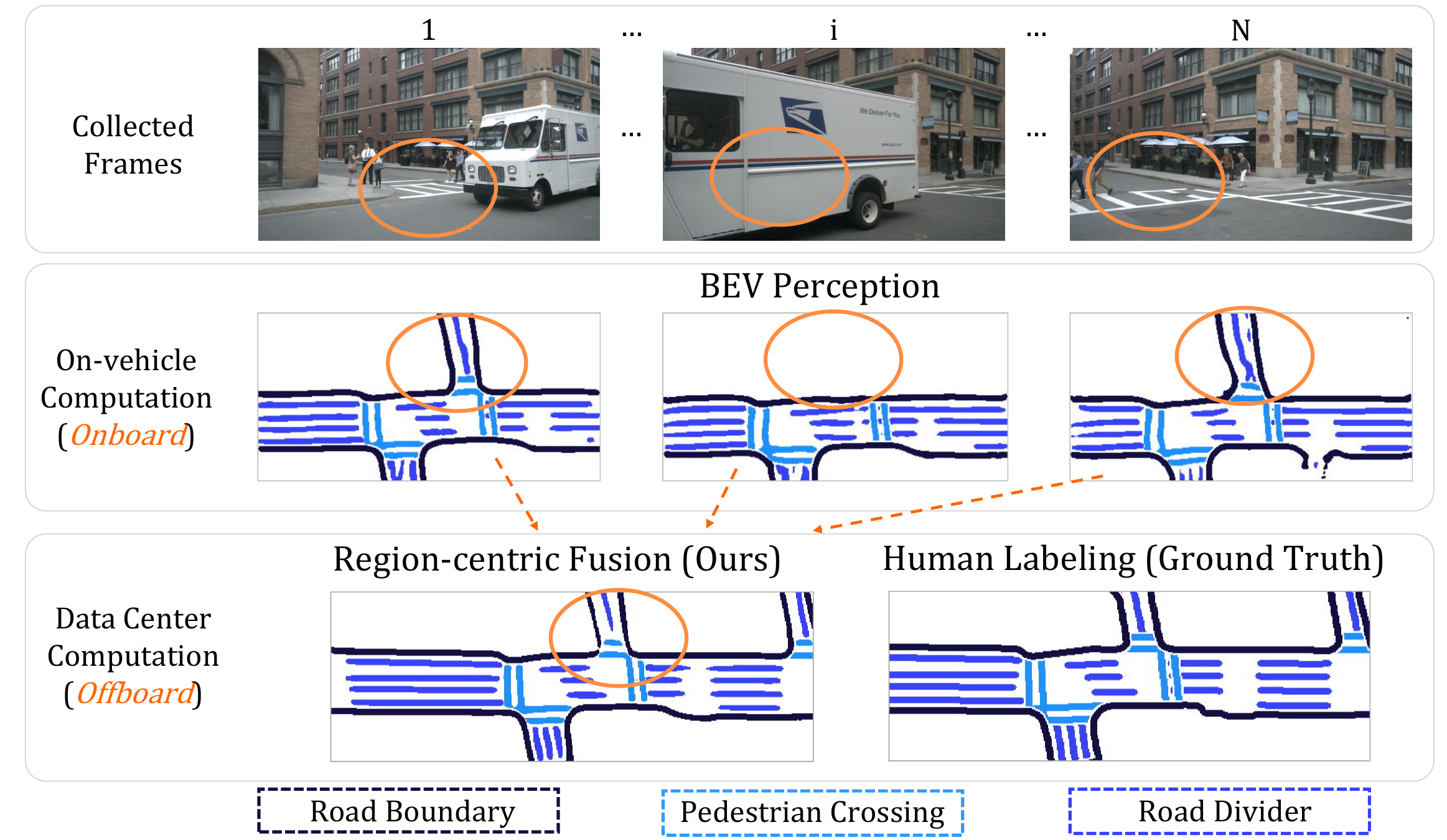}
    \caption{Current \emph{onboard} methods generate unreliable HD map predictions that are inconsistent across multiple views, due to occlusions or viewpoint changes. By contrast, our \emph{offboard} pipeline constructs a unified and multi-view consistent HD map with clearer lanes. Our key design is a \emph{region-centric} framework that aggregates single-frame information for each target HD map region.}
    \label{fig:teaser_multi_view}
    \vspace{-4mm}
\end{figure}

High-definition maps (HD maps) play a crucial role in ensuring the safe navigation of autonomous vehicles, by providing essential positional and semantic information about road elements. Ideally, one would expect the process of constructing HD maps to be as simple as collecting numerous sensory data while driving and then utilizing an \emph{automatic} algorithm to extract the road elements, as illustrated in Fig.~\ref{fig:teaser_multi_view}. However, the mainstream solutions generally involve human annotators, as seen in widely-used datasets~\cite{caesar2020nuscenes, chang2019argoverse, ettinger2021large, wilson2023argoverse}. This design is based on the consideration of the infrastructure role and high re-usability of HD maps, which can serve autonomous vehicles for virtually \emph{infinite} times after a \emph{single} construction process before any environmental changes. Even so, the expense of manual annotation obstructs the expansion of autonomous driving to new locations, and we aim to develop reliable algorithms that can decrease or replace the need for human labor in HD map construction. 

Towards this goal, there have been recent attempts that automatically generate HD maps using bird's-eye-view (BEV) perception~\cite{ harley2022simple, li2022hdmapnet, li2022bevformer}. However, their results are often unreliable, as illustrated by noticeable inconsistencies in the predicted HD maps from different viewpoints (a representative example is in Fig.~\ref{fig:teaser_multi_view}). We argue that \emph{multi-view consistency is an intrinsic property of HD maps}, because the rigid and static HD maps shouldn't change significantly after simply switching viewpoints. The violations of consistency arise from the fact that existing BEV perception algorithms do not account for all the views explicitly and thus do not align their predictions. This issue further boils down to their \emph{onboard} setting, where the models are only allowed to access computing devices \emph{onboard} in autonomous vehicles and can only handle a single frame or a few neighboring frames.

Given such limitations of the \emph{onboard} setup, we underline a critical yet under-explored \emph{offboard} setup that removes the computation constraints. Our offboard setting aligns well with the \emph{infrastructure} role of HD maps: constructing HD maps can and should utilize powerful data centers to maximize the fidelity of predictions, thus ensuring the safety and reliability of the virtually infinite usages of HD maps. By aggregating information from diverse viewpoints and enhancing consistency, our offboard generation provides a natural improvement. As shown in Fig.~\ref{fig:teaser_multi_view}, having multiple views of a shared region offers richer geometric and semantic cues, as well as improves the completeness of scene understanding, particularly regarding frequent occlusions in urban traffic.

\begin{figure}
    \centering
    \includegraphics[width=1.0\linewidth]{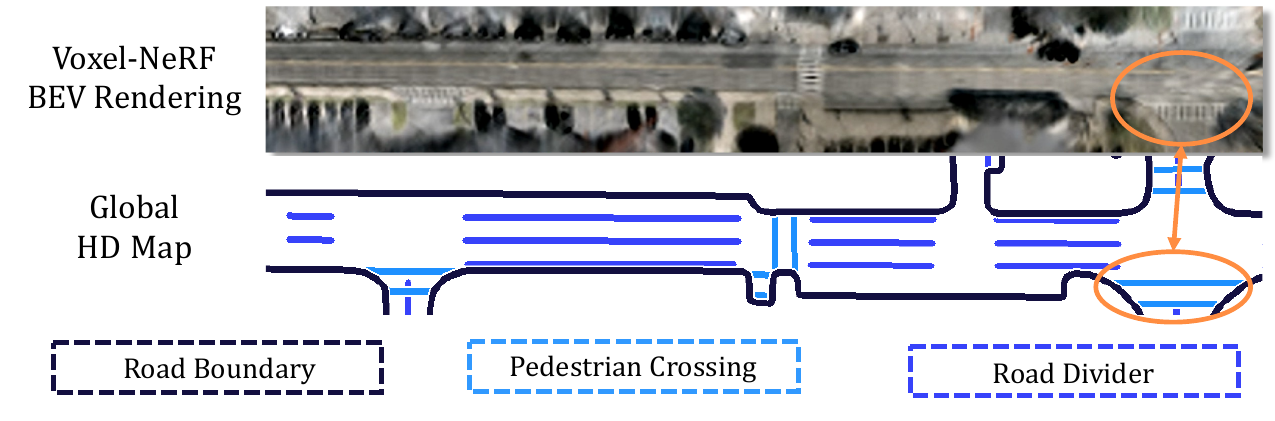}
    \vspace{-4mm}
    \caption{Our Voxel-NeRF reconstructs high-quality 3D structure of the scene. As in the rendering result, the lanes and pedestrian crossings (highlighted) are clear.}
    \label{fig:intro_nerf_show}
    \vspace{-4mm}
\end{figure}

With HD map construction primarily relying on vision information to infer the semantics, which is different from the previous offboard studies depending on point clouds, we are the first to explore a vision-oriented offboard perception framework to our best knowledge. To this end, we propose {\em \ourmodelfull (\ourmodel)} that leverages information from every frame's viewpoint and generates a unified HD map consistent with all of them. In contrast to the \emph{frame-centric} design in current onboard methods that merges a fixed number of frames on the input level, we propose a \emph{region-centric} design inspired by ``offboard 3D detection''~\cite{qi2021offboard} to fully utilize the data from diverse views. Notably, our design can connect every HD map region with an arbitrary number of input frames covering its area. The pipeline of our framework involves extracting all the HD map patches predicted by an off-the-shelf onboard model related to that HD map region, and then fusing the patches into a final result that agrees with all the views, as illustrated in the arrows in Fig.~\ref{fig:teaser_multi_view}. To give more weight to reliable frames, such as those with a clear view of the target region, we introduce an ``\emph{uncertainty network}'' as a key component, which assigns confidence scores to onboard results and performs a weighted average of HD map patches guided by the confidence.

We further enhance the consensus among all the frames by augmenting the uncertainty network with cross-view consistency information. Our key insight is to learn a coherent 3D structure from diverse views and provide it as an auxiliary input to the uncertainty network. For this purpose, we exploit neural radiance fields (NeRFs)~\cite{nerf}, a state-of-the-art approach that represents 3D structures of scenes. As shown in Fig.~\ref{fig:intro_nerf_show}, our NeRF model synthesizes a high-quality scene structure. Compared with alternative 3D reconstruction strategies like structure from motion (\eg, COLMAP~\cite{schoenberger2016sfm}), NeRF is more preferred from a practical perspective, because its runtime grows linearly with the frame number, whereas COLMAP increases quadratically. Moreover, NeRF is {\em fully self-supervised} and does not require additional annotations, unlike multi-view stereo methods such as MVSNet~\cite{yao2018mvsnet}. To improve the scalability of NeRF, we leverage a voxelized variant of NeRF (Voxel-NeRF) to promote efficiency and propose loss functions that implicitly guide the concentration of NeRF on the near-ground geometry related to HD map generation. Additionally, we highlight NeRF's flexibility and scalability to an arbitrary number of views, making it critical in offboard HD map generation.

To summarize, we make the following contributions:
\begin{enumerate}[leftmargin=*, noitemsep, nolistsep]
    \item We are the {\em first} to study the problem of learning to generate HD maps \emph{offboard}, and we are also the first \emph{vision-oriented} offboard study to our best knowledge.
    \item We propose an effective \emph{region-centric} framework \ourmodel that can generate a multi-view consistent HD map from an arbitrarily large number of frames. 
    \item We introduce and extend Voxel-NeRF to encode the 3D structure from all frames for HD map generation tasks, further guiding the fusion for multi-view consistency.
\end{enumerate}
Large-scale experiments on nuScenes~\cite{caesar2020nuscenes} show that \ourmodel significantly improves HD map quality. Notably, \ourmodel can effectively utilize an increasing number of input frames, making it attractive for real-world applications.
\section{Related Work}
\mypar{Offboard 3D perception.}
The need for large-volume training data encourages developing offboard algorithms. Existing studies mainly focus on predicting 3D bounding boxes~\cite{najibi2022motion, pang2021model, qi2021offboard, yang2021auto4d}. The most representative work on ``offboard 3D detection''~\cite{qi2021offboard} extracts multi-frame point clouds in object tracks and refines the 3D bounding boxes with the ``4D'' data. Its success heavily relies on the absolute 3D positions of point clouds, where simply overlaying LiDAR points can construct denser surfaces of objects. However, in HD map generation that relies on images, it is not straightforward to accumulate imagery data in the 3D space. To overcome this limitation, we propose region-centric fusion to aggregate multi-frame information and utilize multi-view reconstruction, \emph{e.g.} NeRF, to encode global geometry. Our study is also the \emph{first vision-oriented offboard} pipeline.

\mypar{BEV segmentation and HD map construction.} Onboard HD map construction is closely related to BEV segmentation, as in HDMapNet~\cite{li2022hdmapnet}. The major challenge in BEV segmentation is to map the image features to the 3D world. The conventional approach leverages inverse perspective warping~\cite{ipm-02,ipm-01,ipm-03,vpn,ipm-04}. BEV perception methods either apply attention to capture the transformation~\cite{li2022bevformer, bevfusion}, incorporate depth information~\cite{FIERY, bevdepth, bevseg, lss}, or directly query the features from voxels~\cite{harley2022simple}. To better support downstream applications, some recent methods~\cite{liao2022maptr, liu2022vectormapnet} have developed special decoders to generate vectorized HD maps. Unlike these \emph{onboard} methods, our proposition is a general \emph{offboard} pipeline that utilizes any off-the-shelf segmentation models as an internal component and refines its results with multi-view consistent fusion. In this sense, neural map prior~\cite{xiong2023neural} and NeMO~\cite{zhu2023nemo} also propose to perform long-term temporal fusion, but they primarily focus on an onboard setting and cannot fully leverage the multi-view consistency from long video sequences. Compared with them, our \ourmodel is an offboard framework and explicitly accounts for the geometry of diverse views by reconstructing the 3D structure of the scene.

\mypar{Neural radiance fields.} NeRF~\cite{nerf} has shown outstanding capability in 3D reconstruction. Recent work \cite{nerfW, urban-nerf, blocknerf, xie2023snerf} has extended NeRF into large unbounded scenes, such as city-scale NeRF with ego-centric camera settings~\cite{urban-nerf, blocknerf, xie2023snerf} and improvement from depth-supervised methods~\cite{deng2021depth, donerf, nerfmvs, pointnerf}. With NeRF's ability to optimize 3D structures from numerous views, it becomes an ideal method to enforce multi-view consistency for offboard perception. However, as we are the first to adapt NeRF for HD map generation, some important modifications are made.  First, we adopt voxel-based NeRF \cite{plenoxels, nsvf, instant-ngp, dvgo, dvgo-v2} to accelerate the NeRF training by voxelizing the space and encoding the parameters for each position in the voxels. This allows us to reconstruct a huge scene from nuScenes within minutes. In addition, we propose a ``total-variance loss'' to enhance NeRF's concentration on the near-ground geometry, which also reflects the shift of concentration from pixel quality to downstream HD map generation. 
\begin{figure*}[tb]
    \centering
    \includegraphics[width=0.85\linewidth]{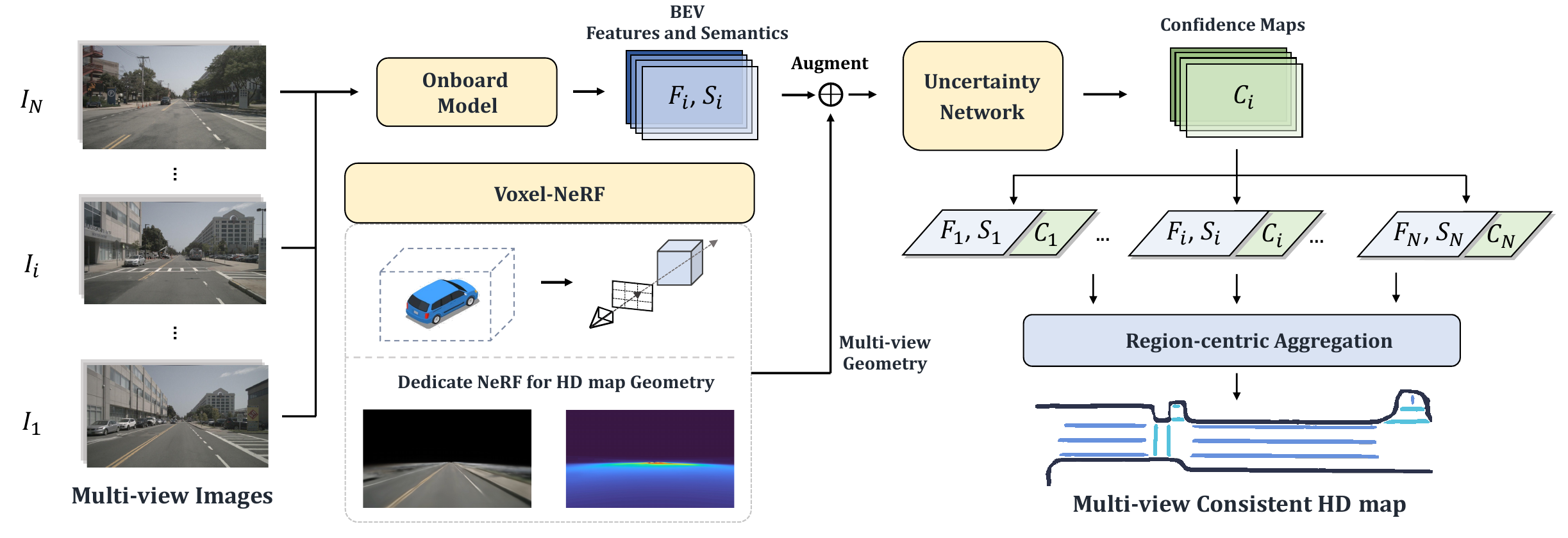}
    \vspace{-2mm}
    \caption{Offboard pipeline of \ourmodel. Given an arbitrary number of input frames, \ourmodel first leverages an off-the-shelf \emph{onboard model} to generate BEV features and semantic maps for each frame. Then an \emph{uncertainty network} predicts their corresponding confidence maps and guides the region-centric aggregation of a unified HD map. Our pipeline further develops a Voxel-NeRF tailored to 3D structures related to HD maps to augment \ourmodel with multi-view geometry. }
    \vspace{-3mm}
    \label{fig:framework}
\end{figure*}

\section{Offboard HD Map Generation}
\label{sec:offboard}

Given a sequence of sensory data, the goal of HD map generation is to predict the positions and semantics of road elements in the BEV space, including road dividers, road boundaries, and pedestrian crossings. 

\mypar{Problem statement.}  We consider the input of HD map generation as $\mathcal{D}=\{(I_i, P_i)\}_{i=1}^N$, where $I_i$ denotes the $i$-th sensor frame, $P_i$ is the set of associated sensor poses, and $N$ is the total number of frames in the database representing diverse views of a scene captured by a moving ego vehicle. The output is denoted as $\mathcal{M}=\{M_i\}_{i=1}^N$, where $M_i$ is the HD map for the region nearby the ego vehicle on frame $i$. Following HDMapNet~\cite{li2022hdmapnet}, we define $M_i$ as a local semantic map on BEV. Note that the aforementioned formulation is agnostic to sensor types. In the main paper, we mainly focus on {\em vision-oriented} HD map generation, and we extend it to leveraging additional LiDAR data in Sec.~\ref{sec:lidar}. Specifically, every frame $I_i$ contains $K=6$ RGB images $\{I_{i,j}\}_{j=1}^K$ on nuScenes~\cite{caesar2020nuscenes}, and $P_i=\{P_{i,j}\}_{j=1}^K$ comprises of the intrinsic and extrinsic matrices of corresponding cameras.

\mypar{Offboard vs. onboard settings.} Compared with the conventional onboard setup, our offboard setup offers greater flexibility in terms of speed and computation resources. Onboard HD map generation algorithms are often constrained by efficiency requirements and cannot use all the $N$ frames {\em in a single run}. By contrast, offboard algorithms are allowed to have access to all the $N$ frames, and can then leverage the offline setting and abundant computation resources to generate HD maps of higher quality.

\mypar{From \emph{frame-centric} to \emph{region-centric} designs.} There are different strategies to utilize the temporal data from $N$ frames, similar to offboard 3D detection~\cite{qi2021offboard}. A direct solution is \emph{frame-centric}~\cite{qi2021offboard}, in which we na\"ively increase the number of frames for existing \emph{onboard} HD map construction methods, typically BEV segmentation models, and extend them to long sequences. While previous work~\cite{li2022hdmapnet, li2022bevformer} has illustrated the benefit of longer temporal horizons, a multi-frame BEV segmentation model can only handle a fixed number of input frames, and increasing the frame number requires a linear growth in GPU capacity. Therefore, simply scaling up the input frames of existing onboard models is not an effective way of exploiting the offboard data, which often have varying and large frame numbers.

To overcome the limitations of the frame-centric design, we propose a novel \emph{region-centric} design that adaptively aggregates information from an arbitrary number of available frames for each HD map region. Our design is inspired by the \emph{object-centric} notion in 3D detection~\cite{qi2021offboard}, but extends to the task of HD map construction. Doing so enables the consensus across frames captured from different viewpoints.

\section{Method: \ourmodelfull}
\label{sec:method}

\mypar{Overview.} Fig.~\ref{fig:framework} illustrates the overall framework of our \ourmodelfull (\ourmodel). An onboard HD map model processes every frame $(I_i, P_i)$ and generates its corresponding BEV feature map $F_i$ and HD map semantics $S_i$ (Sec.~\ref{sec:onboard}). Then, an uncertainty network assesses the reliability of the single-frame information $F_i$ for every region on the HD map (Sec.~\ref{sec:uncertainty}). Meanwhile, a voxelized NeRF $f_{\text{NeRF}}$ optimizes a global 3D structure from all $N$ frames and provides multi-view consistency information to the uncertainty network (Sec.~\ref{sec:voxel_nerf}). The final prediction for every region on the HD map is produced by a weighted average of the single-frame semantics $S_i$, which enables handling an arbitrary number of frames.

\subsection{Onboard Model}
\label{sec:onboard}

The onboard model is the entry point of our pipeline. Most existing HD map generation methods follow an encoder-decoder design. The encoder generates a BEV feature map $F_i$ from the input $(I_i, P_i)$ as $\mathtt{Encoder}(I_i, P_i)\xrightarrow{} F_i$, and the decoder converts the feature map $F_i$ into a semantic map $S_i$ as $\mathtt{Decoder}(F_i)\xrightarrow{}S_i$.

Since our pipeline only requires the BEV feature map $F_i$ to activate the subsequent modules, \ourmodel is agnostic to specific encoder-decoder designs. Without the loss of generality, we mainly adopt the encoder in SimpleBEV~\cite{harley2022simple} and use a lightweight convolutional decoder. Results based on additional models are in Table~\ref{supsec:auto_labeling} (Supplementary).

\mypar{Encoder.} For each frame, a convolutional backbone first converts $K$ images $\{I_{i,j}\}_{j=1}^K$ into 2D image feature maps $\{F_{i,j}^\mathrm{2D}\}_{j=1}^K$. The features are then {\em lifted} into the 3D world, through a set of voxels that are pre-defined by the encoder with shape $X\times Y\times Z$ centered around the ego vehicle: the 2D features are bi-linearly sampled for every voxel based on their projected locations on the images, leading to a voxelized 3D feature map $F_{i}^\mathrm{3D}$. Finally, reducing the Z-axis of $F_{i}^\mathrm{3D}$ produces a BEV feature map $F_i$ with shape $X\times Y\times C$, where $C$ is the feature dimension.

\mypar{Decoder.} Our decoder is a fully-convolutional segmentation head that predicts the logits of semantics from every BEV grid in $F_i$. It generates the surrounding HD map as the semantic segmentation result $S_i$ 
 with shape $X\times Y$.

\mypar{Region-centric extension.} Our \emph{region-centric} design considers each BEV grid as an HD map region. If a grid is covered in $N'$ frames, it receives $N'$ features and predictions from different viewpoints. \ourmodel then fuses information from $N'$ view-specific frames to create a multi-view consistent feature for this region, detailed as below.
\vspace{-1mm}
\subsection{Global Aggregation via Uncertainty Network}
\label{sec:uncertainty}
\begin{figure}
    \includegraphics[width=0.9\linewidth]{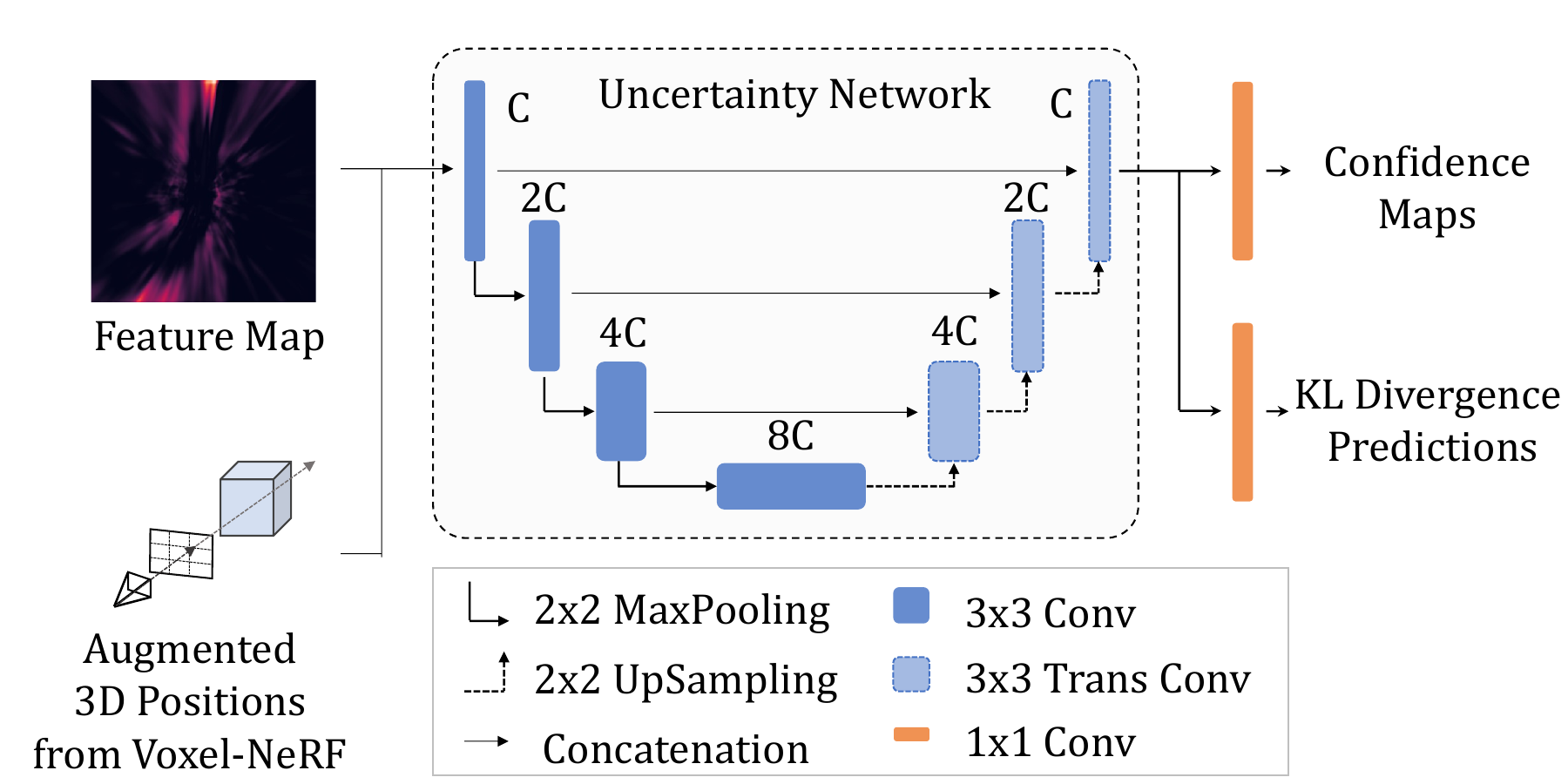}\vspace{-2mm}
    \caption{Architecture of {\bf uncertainty network}, which takes as input the feature map and the augmented 3D positions from NeRF for each voxel (Sec.~\ref{sec:voxel_nerf}). It outputs the confidence maps for region-centric fusion and, optionally, the predicted KL-divergence for the KL-divergence loss (detailed in Sec.~\ref{sec:uncertainty}).}
    \label{fig:uncetainty-net}
    \vspace{-4mm}
\end{figure}

\vspace{-1mm}
\mypar{Region-centric uncertainty-aware fusion.} Our region-centric offboard pipeline learns to aggregate the $N$ frames of independent HD map predictions $\{S_i\}_{i=1}^N$ into a multi-view consistent prediction for each region. Our key design is to introduce an \emph{uncertainty network}. For the HD map predictions from all viewpoints, the uncertainty network assigns a confidence score to each BEV grid, resulting in $N\times X\times Y$ scores that reflect the pairwise reliability of a viewpoint contributing to an HD map region. Specifically, the uncertainty network takes the BEV features $\{F_i\}_{i=1}^N$ as input and generates the confidence maps $\{C_i\}_{i=1}^N$, with $C_i$ of shape $X\times Y$. In Sec.~\ref{sec:voxel_nerf}, we will describe how we further incorporate global geometry encoded by Voxel-NeRF into the uncertainty network. The architecture of our uncertainty network is illustrated in Fig.~\ref{fig:uncetainty-net}, adopting a UNet-like~\cite{ronneberger2015unet} structure for predicting densely on every voxel.

We then aggregate the per-frame semantics and confidences into a final HD map. Suppose an arbitrary target position $(x^w, y^w)$ is specified in the world coordinate system, we transform it to the local coordinate system of every frame with poses $\{P_i\}_{i=1}^N$ and sample the semantic maps and confidence scores at corresponding locations. Finally, the prediction for $(x^w, y^w)$ is obtained by a weighted average of per-frame semantics according to their confidences.

\mypar{KL-divergence loss for enhanced uncertainty learning.} In addition to generating confidence scores, we add a multi-layer perceptron (MLP) head to the uncertainty network to infer the KL-divergence between the predicted and ground truth semantics, as shown in Fig.~\ref{fig:uncetainty-net}. Intuitively, learning to regress the KL-divergence value between the predicted and ground truth HD maps augments the confidence score of the uncertainty network, because a smaller KL-divergence value indicates better quality of map construction and the uncertainty network should assign higher confidence accordingly. During training, we encourage the inferred divergence $\text{KL}^U$ to be close to the true divergence $\text{KL}^G$ between semantics $S_i$ and $S_i$'s ground truth, formally,
\begin{align}
\label{eqn:kl_loss}
&\mathcal{L}_\text{KL} = \frac{1}{XY}\sum_{x=1}^{X}\sum_{y=1}^{Y}\|\text{KL}^G[x, y]-\text{KL}^U[x, y]\|_2^2.
\end{align}

We train the uncertainty network with both a cross-entropy loss between the fusion result and the ground truth semantics at each location $(x^w, y^w)$~\cite{li2022hdmapnet} and our auxiliary KL-divergence loss in Eqn.~\ref{eqn:kl_loss}. Given that the weighted average operation is differentiable, the gradients from both of the loss terms can be back-propagated to the confidence scores for updating the uncertainty network.

\vspace{-1mm}
\subsection{Voxel-NeRF for Multi-view Consistency}
\label{sec:voxel_nerf}
\vspace{-1mm}
\ourmodel further leverages a voxelized NeRF to effectively construct a {\em unified} 3D structure of the scene from the $N$ frames, which is incorporated with the uncertainty network to improve the multi-view consistency of HD maps.  

\mypar{Voxel-NeRF for traffic scenes.} 
NeRF~\cite{nerf} represents a 3D scene as a continuous function $f_{\text{NeRF}}: (\mathbf{x}, \mathbf{\theta}) \rightarrow (\mathbf{c}, \sigma)$, which maps every point $x$ in the 3D space to its color $c$ and density $\sigma$, relative to the viewing direction $\theta$. By explicitly encoding camera projection in the neural rendering process, the learned NeRF model $f_{\text{NeRF}}$ encodes the 3D geometry of the corresponding scene from input images. Despite the success of the vanilla NeRF, applying it to autonomous driving datasets poses significant challenges, because of the unbounded nature of the scenes and the huge quantities of data involved (\eg, 850 scenes on nuScenes~\cite{caesar2020nuscenes}). Therefore, we introduce a voxelized NeRF based on DVGO~\cite{dvgo-v2} for better training speed and scalability. Our Voxel-NeRF captures multi-view consistent geometry for outdoor scenes, instead of small objects as in conventional NeRFs. To achieve this, we initialize voxel grids with shape $X_s\times Y_s\times Z_s$ to cover the entire scene, which is larger than $X\times Y\times Z$ used in onboard models for single-frame areas. For each camera ray, the neural rendering operation in Voxel-NeRF \emph{concurrently} queries every voxel intersected by the ray. This concurrent querying of voxels significantly accelerates the training of our Voxel-NeRF, {\em from hours to minutes} for any scene in nuScenes, enabling a reasonable computation budget for \ourmodel. More details can be found in Sec.~\ref{sec:implementation_details}.

\mypar{Augmenting uncertainty network.} Conceptually, the predicted semantics at a position is more reliable when it resides on the object surfaces. Once NeRF produces a multi-view consistent structure, we can compute the distance between each voxel center and its closest surface. Such a clue can be exploited to evaluate the reliability of semantic maps. To this end, for an arbitrary $(x, y)$ on BEV, we first recover all the voxel center locations at the BEV coordinate $(x, y)$ as $L_{\text{Voxel}} = \{(x, y, z_i)\}_{i=1}^Z$ and then compute their corresponding pixel locations on the images $L_{\text{Image}}=\{(x^p_i, y^p_i\}_{i=1}^Z$. By volume rendering along the camera rays crossing these pixels, $f_{\text{NeRF}}$ reconstructs the 3D positions of these pixels, denoted as $L_{\text{NeRF}}=\{(x_i^R, y_i^R, z^R_i)\}_{i=1}^Z$, which are generally the intersections between their camera rays and surfaces. (Ray casting details are in Sec.~\ref{supsec:ray_casting} (Supplementary).) By calculating $\Delta{L}_{\text{N/V}}=\{(x_i^R - x, y_i^R - y, z^R_i - z_i)\}_{i=1}^Z$, we assess the consistency between voxel centers and the global 3D structure. Finally, we employ an MLP upon $\Delta{L}_{\text{N/V}}$ and concatenate its output with the BEV feature, which is then used as the augmented input to the uncertainty network, which is the ``augmented 3D positions'' in Fig.~\ref{fig:uncetainty-net}.

\begin{table*}[t]
\caption{Comparison with state-of-the-art vision-based HD map generation methods on nuScenes~\cite{caesar2020nuscenes}. ``*'' means the results reported in HDMapNet~\cite{li2022hdmapnet}. ``Average Fusion'' is an offboard baseline explained in Sec.~\ref{sec:comp_sota}. The quantitative results indicate that our \ourmodel has significant benefits to HD map generation and outperforms offboard baseline approaches.}
\vspace{-2mm}
    \centering
    \resizebox{0.70\linewidth}{!}{
    \begin{tabular}{l|l|ccccc}
    \toprule
    \multirow{2}{*}{Setup} & \multirow{2}{*}{Method}  & \multicolumn{4}{c}{mIoU (Short-range / Long-range)} \\ & & Divider & Ped Crossing & Boundary & All  \\ \midrule
    \multirow{6}{*}{Onboard} & IPM(B)*  & 25.5 /\;\;\;-\;\;\;    & 12.1 /\;\;\;-\;\;\;   & 27.1 /\;\;\;-\;\;\;   & 21.6 /\;\;\;-\;\;\; \\
    & IPM(B+C)*           & 38.6 /\;\;\;-\;\;\;    & 19.3 /\;\;\;-\;\;\;   & 39.3 /\;\;\;-\;\;\;   & 32.4 /\;\;\;-\;\;\; \\
    & VPN*                & 36.5 /\;\;\;-\;\;\;    & 15.8 /\;\;\;-\;\;\;   & 35.6 /\;\;\;-\;\;\;   & 29.3 /\;\;\;-\;\;\; \\
    & Lift-Splat-Shoot~\cite{lss}*   & 38.3 /\;\;\;-\;\;\;    & 14.9 /\;\;\;-\;\;\; & 39.3 /\;\;\;-\;\;\;   & 30.8 /\;\;\;-\;\;\; \\
    & HDMapNet \cite{li2022hdmapnet}    & 40.6 / 33.9  & 18.7 / 19.4  & 39.5 / 34.9 & 32.9 / 29.4  \\
    & Onboard Model (Ours) & 46.4 / 39.3 & 29.7 / 26.4 & 48.1 / 39.1 & 41.4 / 35.0  \\
    \midrule
    \multirow{2}{*}{Offboard} & 
     Average Fusion &  
     48.86 / 42.83 & 
     31.55 / 24.75&
     51.98 / 43.91 &
     44.13 / 37.16 \\
     & \ourmodel (Ours) & 
     \textbf{50.87} / \textbf{48.15} & 
     \textbf{34.52} / \textbf{33.34} & 
     \textbf{55.64} / \textbf{50.28} & 
     \textbf{47.01} / \textbf{43.92}  \\
    \bottomrule
    \end{tabular}
    }
    \label{tb-all}
    \vspace{-4mm}
\end{table*}

\mypar{Dedicating NeRF for HD maps with total-variance loss.}
Note that {\em our objective is to facilitate HD map generation, rather than optimizing rendering quality}. With the majority of HD map elements situated on the ground, we modify the NeRF to focus less on the quality of pixels in the air. To this end, we introduce a simple yet effective ``\emph{total-variance loss}'' that guides the optimization of near-ground geometry \emph{implicitly}. This total-variance loss $\mathcal{L}_{\text{TV}}$ is obtained by accumulating the total-variance $\text{TV}(\cdot)$ at each BEV position:
\begin{align}
    \label{eqn:tv_loss}
    \mathcal{L}_{\text{TV}} = -\frac{1}{X_s Y_s}\sum_{x=1}^{X_s}\sum_{y=1}^{Y_s}\text{TV}(x, y).
\end{align}
Here the total-variance $\text{TV}(\cdot)$ is defined as the L2-norm of the differences of occupancies along the Z-axis, given by
\begin{align}
    \label{eqn:tv}
    \text{TV}(x, y) = \left\|O[x, y, 2\!:\!Z_s]-O[x, y, 1\!:\!Z_s-1]\right\|_2,
\end{align}
where $O[x, y, z]$ represents the density of voxel $(x, y, z)$ predicted by NeRF and $\|\cdot\|_2$ denotes the L2-norm.

We emphasize the ``negative'' sign in Eqn.~\ref{eqn:tv_loss}. It indicates ``maximizing'' the variance, because an accurate ground plane has a \emph{peak} distribution of voxel occupancy on the Z-axis instead of a \emph{uniform} one. TV-loss enables Voxel-NeRF to assign larger densities to the ground plane than transient objects, leading to high-quality 3D structures as in Fig.~\ref{fig:intro_nerf_show}. 
\vspace{-1mm}
\subsection{Training and Inference}
\label{sec:train_and_inference}
\vspace{-1mm}
The procedure of our offboard pipeline follows three steps: (1) we adopt an existing onboard model, (2) train Voxel-NeRF on sequences, and (3) train and infer the uncertainty network. We describe these steps in order and leave detailed configurations in Sec.~\ref{supsec:implementation} (Supplementary).

\mypar{Onboard model.} As \ourmodel is agnostic to the choice of onboard models (Sec.~\ref{sec:onboard}), here we adopt an \emph{off-the-shelf} BEV segmentation model and freeze its parameters during both training and inference stages of the offboard pipeline.

\mypar{Voxel-NeRF.} We train the Voxel-NeRF for all sequences in our training and validation datasets, using both conventional photometric loss and our total-variance loss (Sec.~\ref{sec:voxel_nerf}). Note that our NeRF training is entirely \emph{self-supervised and does not require any annotations}. 

\mypar{Uncertainty network.} The \emph{region-centric} design enables the uncertainty network to handle varying frame numbers of offboard data. In practice, however, the GPU capacities and batching during training limit the network to a fixed and restricted frame number. To overcome this issue, we adopt the solution from video-based tasks (\eg, 2D multi-object tracking~\cite{meinhardt2022trackformer, zeng2022motr}), where models are trained on \emph{short video clips} but are inferred iteratively on \emph{unbounded sequences}. 

Similarly, given the input $N$ frames of a scene, the uncertainty network is trained with samples containing $M (M < N)$ adjacent frames to fit into limited GPU memory. The loss is a weighted sum of our KL-divergence loss and a BEV segmentation loss (Sec.~\ref{sec:uncertainty}). During inference, we apply the uncertainty network to all the $N$ frames independently and use region-centric aggregation to fuse single-frame semantics into a unified HD map.
\vspace{-1mm}
\section{Experiments}
\label{sec:implementation_details}
\vspace{-1mm}
\subsection{Dataset and Implementation Details}
\vspace{-1mm}
\mypar{Dataset.} We conduct experiments on a large-scale autonomous driving dataset: nuScenes \cite{caesar2020nuscenes}. It contains 850 videos with 28,130 and 6,019 frames for training and validation, respectively. On each timestamp, six surrounding cameras collect high-resolution images as input. 

\mypar{Evaluation metrics.} Following prior work~\cite{li2022hdmapnet}, we compute the intersection-over-union (IoU) for HD map categories: divider, pedestrian crossing, and road boundaries. To highlight the challenge of predicting a scene-scale HD map, our evaluation adopts both a {\em short-range} setting~\cite{li2022hdmapnet} covering 60m $\times$ 30m and a new {\em long-range} setting covering 100m $\times$ 100m, which aligns with the common perception range in self-driving~\cite{dong2022SuperFusion, lue2023sst}. Without further mentioning, we conduct our ablation studies under the more challenging long-range setting.

\mypar{Implementation details.} We follow the training and inference settings in Sec.~\ref{sec:train_and_inference} and discuss the details in Sec.~\ref{supsec:implementation} (Supplementary). We emphasize that \ourmodel is scalable to large volumes of offboard data. Within 15 minutes on a single A40 GPU, our Voxel-NeRF can optimize the 3D structure from each nuScenes sequence, which typically has over 1k images covering regions with an average length of $\sim$300m, less than 1 second per frame. In comparison, the common multi-view stereo baseline of COLMAP~\cite{schoenberger2016sfm} may take several hours or even days. Notably, this is because COLMAP spent most of the time in feature matching, which is pairwise across frames and $\mathcal{O}(N^2)$ regarding the frame number $N$. In comparison, the time complexity of NeRF is $\mathcal{O}(N)$. Moreover, our uncertainty network is trained on the samples with $M=5$ adjacent frames to fit into our GPU memory, but it can jointly handle all the frames ($\sim$40) in a nuScenes sequence during the inference stage, as explained in Sec.~\ref{sec:train_and_inference}.
\vspace{-1mm}
\subsection{Comparison with State-of-the-Art Methods}
\label{sec:comp_sota}
\vspace{-1mm}
As our work represents the \emph{first} study on offboard HD map generation, there are no readily available competing methods. Additionally, our \ourmodel can utilize any off-the-shelf onboard model as its internal component. To ensure a meaningful and fair comparison, we organize the experimental results and analysis in Table~\ref{tb-all} as follows.

First, our onboard model adopts the simple-yet-effective design from SimpleBEV~\cite{harley2022simple}. As shown in the ``onboard'' rows, our onboard model already \emph{consistently} outperforms previous baselines in both short-range and long-range settings. Second, our \ourmodel brings a significant improvement of $\sim$7\% mIoU compared with our already effective onboard model. Notably, our offboard method is better than HDMapNet~\cite{li2022hdmapnet} by around 50\% with over 15\% IoU increase on all the categories. Finally, we develop an offboard baseline algorithm called ``Average Fusion.'' It does not consider the quality of different viewpoints and performs region-centric aggregation by equally averaging the single-frame semantic maps. Compared with ``Average fusion,'' our \ourmodel still improves the HD map quality by a large margin of over $\sim$7\% mIoU under the long-range setting.

\begin{table}[tb]
\caption{\ourmodel significantly improves vectorized HD map quality over both the onboard VectorMapNet~\cite{liu2022vectormapnet} model and the recent temporal fusion method Neural Map Prior~\cite{xiong2023neural} (NMP). ($\dagger$ means the performance reported in NMP; $\ddagger$ means the performance from VectorMapNet's officially released checkpoint.)} \vspace{-2mm}
    \centering
    \resizebox{0.85\linewidth}{!}{
    \begin{tabular}{c|cccc}
    \toprule
    \multirow{2}{*}{Method} & 
    \multicolumn{4}{c}{mAP} \\ & Divider & Ped Crossing & Boundary & All   \\ 
    
    \midrule
    VectorMapNet$^{\dagger}$ &  47.3 &  36.1 &  39.3 & 40.9 \\
    + NMP & 49.6  & 42.9  & 41.9 &  44.8 \\
    \rowcolor{lightgreen}
     $\Delta$ mAP&\textcolor{gray}{+2.3}&\textcolor{gray}{+6.8}&\textcolor{gray}{+2.6}&\textcolor{gray}{+3.9}\\

    \midrule
    
    VectorMapNet$^{\ddagger}$ & 47.7 & 39.8 & 38.9 & 42.1 \\
    + \ourmodel & \textbf{55.0}  & \textbf{46.2}  & \textbf{45.5} & \textbf{48.9} \\
    \rowcolor{lightgreen}
    $\Delta$ mAP &\textcolor{gray}{+\textbf{7.3}}&\textcolor{gray}{+\textbf{6.4}}&\textcolor{gray}{+\textbf{6.6}}&\textcolor{gray}{+\textbf{6.7}}\\
    \bottomrule
    \end{tabular}
}
    \vspace{-5mm}
    \label{tb:vectormap}
\end{table}
\vspace{-1mm}
\subsection{Comparison on Vectorized HD Maps}
\vspace{-1mm}
We further extend our \ourmodel to vectorized HD map generation to demonstrate its wide compatibility. In Table~\ref{tb:vectormap}, we apply \ourmodel to VectorMapNet~\cite{liu2022vectormapnet}, one of the state-of-the-art onboard models generating vectorized HD maps. Specifically, our \ourmodel performs uncertainty-aware fusion on the frozen BEV features of an open-sourced VectorMapNet model. The results show that \ourmodel improves the HD Map generation quality by a large margin of over $\sim$7 mAP, proving its effectiveness in constructing high-quality HD maps. We further compare \ourmodel with the recent neural map prior (NMP)~\cite{xiong2023neural}, which also capitalizes long-term temporal fusion for global HD map generation in Table~\ref{tb:vectormap}.\footnote{NMP is not included for semantic HD map comparison in Sec.~\ref{sec:comp_sota} because it adopts a different resolution for rasterized HD maps.} In the experiments, We adopt the same metric (mAP) as VectorMapNet~\cite{liu2022vectormapnet}. As clearly illustrated, our offboard \ourmodel pipeline is better in both absolute (48.9 mAP vs. 44.8 mAP) and relative improvements (6.8 mAP vs. 3.9 mAP).

\subsection{Ablation Studies}
\label{sec:ablation}

\begin{table}[t]
\caption{The components in \ourmodel effectively improve HD map generation step by step. We analyze the uncertainty network (UN) and KL-divergence loss ($\mathcal{L}_{\text{KL}}$) discussed in Sec.~\ref{sec:uncertainty}, and Voxel-NeRF (NeRF) and total-variance loss ($\mathcal{L}_{\text{TV}}$) discussed in Sec.~\ref{sec:voxel_nerf}.}
\vspace{-2mm}
    \centering
    \resizebox{1.0\linewidth}{!}{
    \setlength{\tabcolsep}{2.2mm}{
    \begin{tabular}{c|cccc|cccc}
    \toprule
    \multirow{2}{*}{ID} & \multicolumn{4}{c|}{Offboard Component} & \multicolumn{4}{c}{mIoU} \\ 
     & UN & $\mathcal{L}_{\text{KL}}$ & NeRF & $\mathcal{L}_{\text{TV}}$ & Divider & Crossing & Boundary & All   \\ 
    \midrule
    1 & \multicolumn{4}{c|}{Onboard Model} & 39.30  & 26.44  & 39.10 & 34.95 \\
    2 & \multicolumn{4}{c|}{Average fusion}       & 42.83  & 24.75 & 43.91 & 37.16 \\
    \midrule
    3 & \cmark &        &        &        & 46.90  & 30.30  & 49.07 & 42.09 \\
    4 & \cmark & \cmark &        &        & 47.38  & 31.11  & 49.53 & 42.67 \\
    5 & \cmark & \cmark & \cmark &        & 47.64  & 32.36  & 49.67 & 43.22 \\
    6 & \cmark &        & \cmark & \cmark & 48.01  & 32.65  & 50.12 & 43.59 \\
    7 & \cmark & \cmark & \cmark & \cmark & \textbf{48.15}  & \textbf{33.34} & \textbf{50.28} & \textbf{43.92} \\
    \bottomrule
    \end{tabular}
    }}
    \vspace{-2mm}
    \label{tb-fusion-strategy}
\end{table}

\mypar{\ourmodel Components.} We quantify the improvement from each offboard module in Table~\ref{tb-fusion-strategy}. (1) \textbf{Region-centric fusion baseline.} Beginning from the onboard model (row 1), we first apply average fusion (row 2) to it (discussed in Sec.~\ref{sec:comp_sota}) as a baseline. The improvement indicates that our region-centric design indeed helps by fusing numerous frames into a unified HD map. (2) \textbf{Uncertainty network.} Replacing the average fusion (row 2) with the uncertainty network (row 3) enables larger contributions from more reliable frames and the $\sim$5\% increase in mIoU validates that assessing quality is critical for better HD map quality. (3) \textbf{KL-divergence loss.} The $\sim$0.5\% mIoU on using KL-divergence loss or not (row 3 and row 4) supports that explicitly supervising the uncertainty network with KL-divergence effectively improves the region-centric fusion to rely on frames with higher quality. (4) \textbf{Voxel-NeRF.} Adding NeRF to a full-fledged uncertainty network further improves the mIoU (row 4 and row 5). In the category-level analysis, we highlight that NeRF is critical for the fusion {\em especially on the challenging structures with smaller regions}, \eg, pedestrian crossings. This evidence validates the importance of global geometry in multi-view consistency. (5) \textbf{Total-variance loss.} Utilizing it further boosts the performance in all the scenarios, validating our effort to dedicate NeRFs for the downstream HD map generation.

\begin{figure}[t]
    \centering
    \setlength{\abovecaptionskip}{0pt}
    \includegraphics[width=0.9\linewidth]{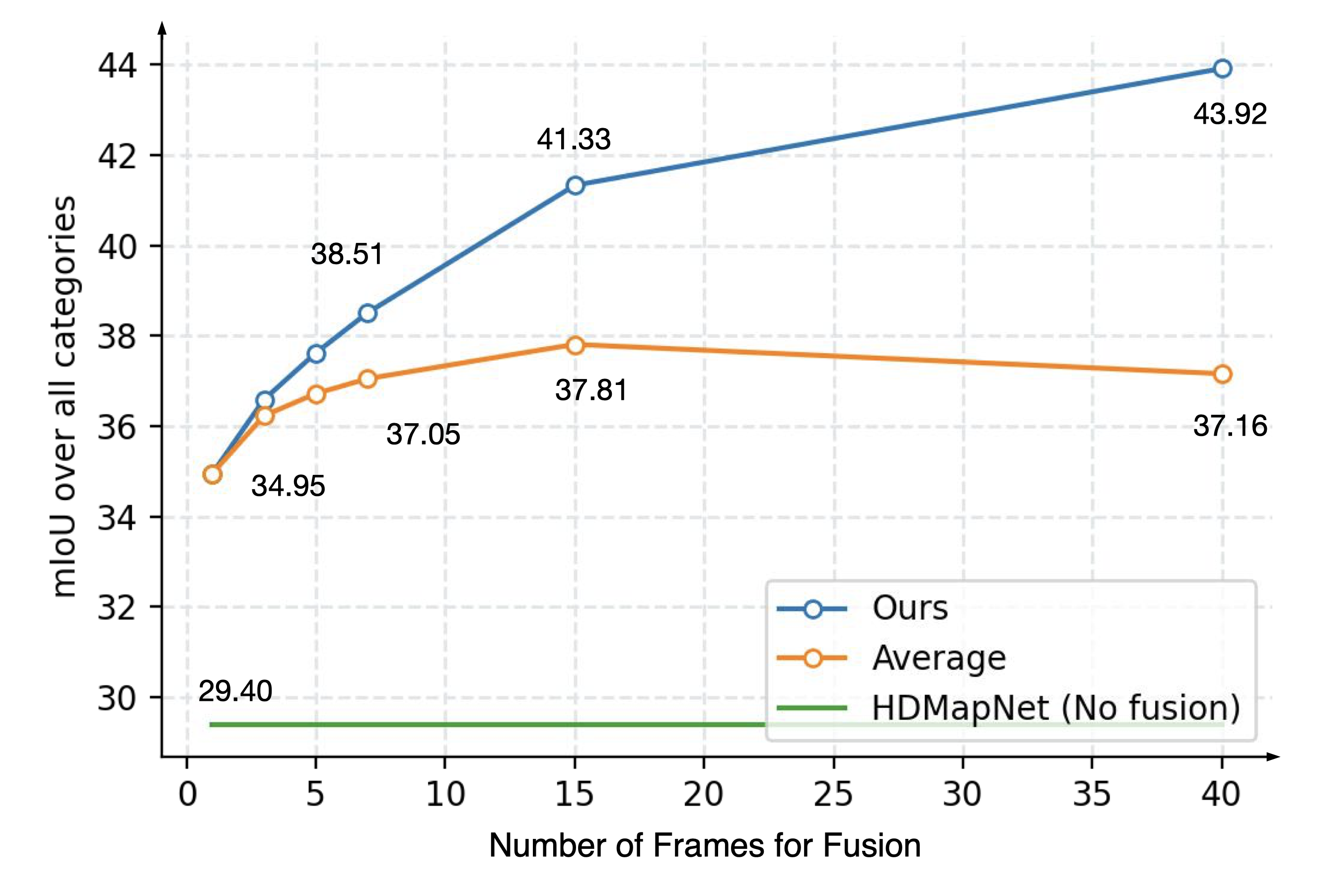}
    \vspace{-2mm}
    \caption{Our \ourmodel can significantly benefit from more input frames, which is attractive for offboard applications. Notably, the performance of the ``average fusion'' baseline saturates and even decreases with more input frames.}
    \label{fig:iou-curve}
    \vspace{-4mm}
\end{figure}

\mypar{Scaling to more frames.} We demonstrate that our fusion strategy can handle and significantly benefit from a larger number of frames, which is critical for offboard HD map generation. We evaluate our offboard framework under varied input frames in Fig.~\ref{fig:iou-curve}. \ourmodel can utilize all the keyframes (40 frames) in nuScenes and this number is only bounded by the sequence length. As clearly shown in the blue curve of Fig.~\ref{fig:iou-curve}, \ourmodel benefits from more frames, indicating its {\em scalability} for offboard scenarios, especially compared with the average fusion baseline, whose performance drops after using more than 15 frames. This indicates that our region-centric fusion strategy is able to {\em reason the complementary regions among the frames}, instead of blindly averaging them all. 

\begin{figure}[tb]
    \centering
    \includegraphics[width=1.00\linewidth]{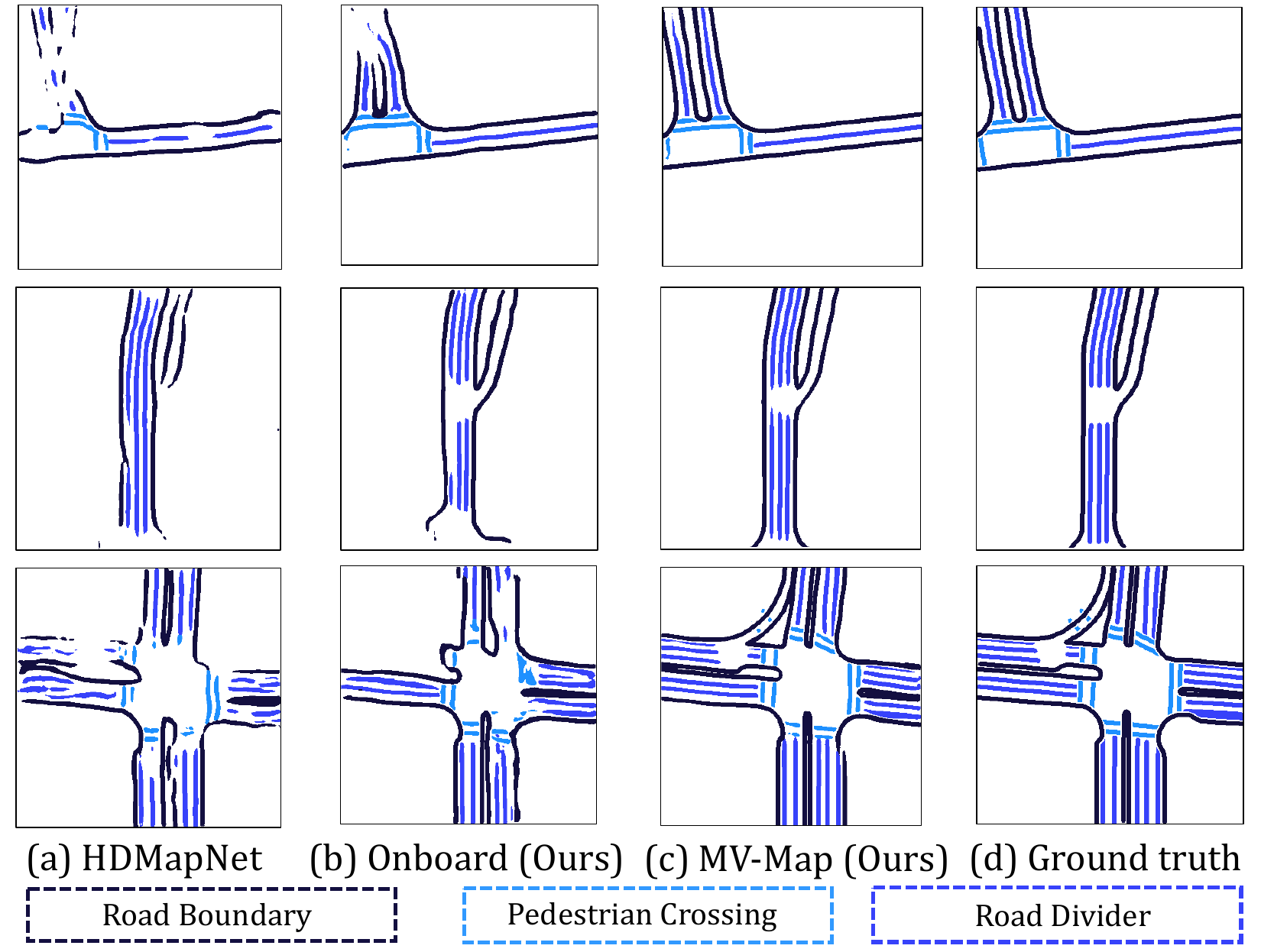}
    \vspace{-3mm}
    \caption{Qualitative comparison in the long-range settings. HD map generated offboard has significantly better quality by fixing the artifacts of the onboard model.}
    \label{fig:cmp_fig}
    \vspace{-4mm}
\end{figure}

\mypar{Qualitative comparison.}  We visualize the generated HD maps in Fig.~\ref{fig:cmp_fig}. As clearly shown, \ourmodel corrects the artifacts from onboard models and achieves better completeness and details. In addition, the HD maps generated offboard have high fidelity compared with the ground truth, especially in the center regions covered by more frames. 

\begin{figure}[tb]
    \centering
    \includegraphics[width=1.0\linewidth]{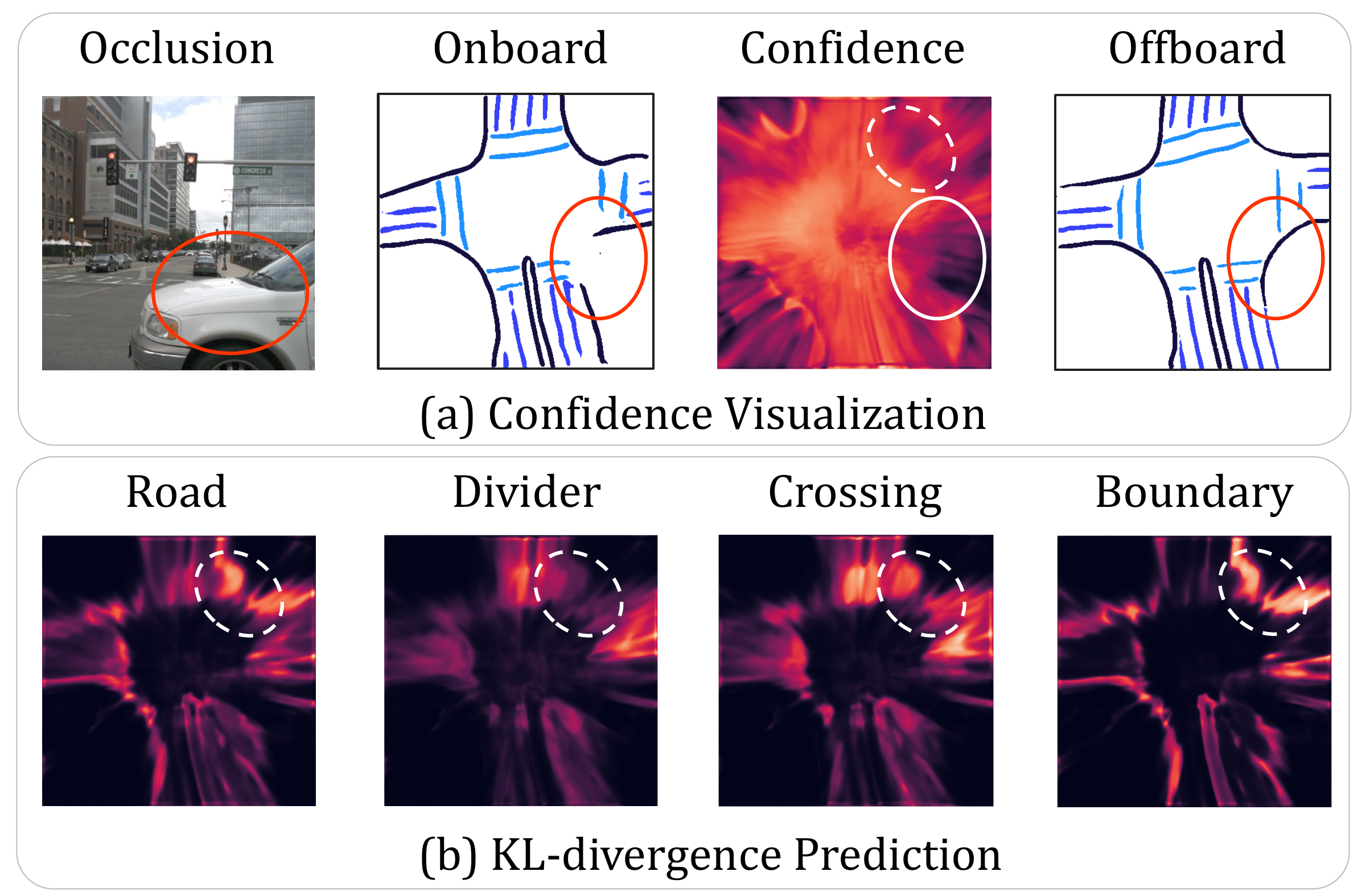}
    \vspace{-4mm}
    \caption{\textbf{(a)} The confidence scores predicted by the uncertainty network \emph{capture the challenging regions} (solid circles) as the occluded region has significantly smaller confidence values than its nearby regions. Darker colors indicate smaller confidence values. \textbf{(b)} Predicted KL divergence between the prediction and ground truth label captures the connection with the predicted confidence scores (dashed circles). The region with a dashed circle has a much larger predicted KL-divergence value than its nearby regions. Accordingly, this area exhibits smaller confidence scores. Darker colors mean smaller KL-divergence values.}
    \label{fig:kl-div}
    \vspace{-4mm}
\end{figure}

\mypar{Analyzing KL-divergence and confidence scores.} We empirically analyze the output of the uncertainty network in Fig.~\ref{fig:kl-div}. As for the confidence scores, we observe that they indeed decrease the contributions of unreliable regions, such as the part with occlusion in Fig.~\ref{fig:kl-div}\textcolor{red}{a}, highlighted with solid circles. The invisible area has much smaller confidence than its nearby regions. Additionally, we transfer the KL-divergence prediction head to the validation set and find the predictions reasonably correlate with the confidence scores, as in Fig.~\ref{fig:kl-div}\textcolor{red}{b}. We notice that the regions with higher KL-divergence values (Fig.~\ref{fig:kl-div}\textcolor{red}{b}) also have lower confidences (Fig.~\ref{fig:kl-div}\textcolor{red}{a}), highlighted with the dashed circles.

\begin{figure*}[ht]
    \centering
    \includegraphics[width=0.8\linewidth]{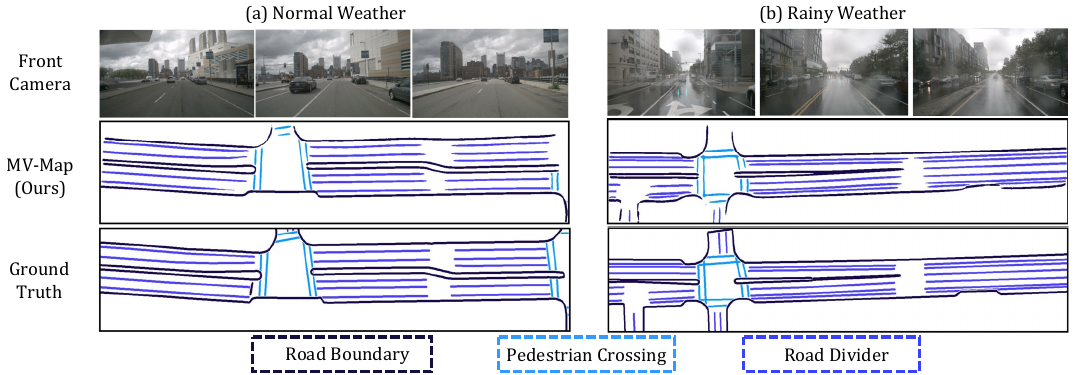}
    \caption{Visualization of a \textbf{unified, scene-scale HD map} through our \ourmodel. It can fuse numerous frames and generate global-scale HD maps with high quality. It is also {\em robust} under different weather conditions and complex road topology.
    }
    \vspace{-3mm}
    \label{fig:global}
\end{figure*} 

\mypar{Using geometric information from data-driven priors.} Our Voxel-NeRF offers geometric information in a {\em fully self-supervised} manner. Meanwhile, our \ourmodel framework is general and can leverage alternative approaches for providing geometric information, such as \emph{learning data-driven priors} from large-scale datasets. We investigate this type of approach here and consider representative monocular depth estimators that are learned off-the-shelf in a {\em supervised} manner. Specifically, we replace the rendering process of Voxel-NeRF with the results from {\em NeWCRFs}~\cite{yuan2022newcrfs} (details in Sec.~\ref{supsec:implementation}, Supplementary). As in Table~\ref{tb:mono-depth}, monocular depth can improve the uncertainty fusion as well (row 2 and row 3). We further notice that NeRF performs slightly better because it \emph{encodes multiple views consistently in a shared 3D structure}, while monocular depth considers each view independently and suffers from scale variation across frames. Encouraged by the benefits of these two distinct types of geometric information, future work is to combine NeRF with learnable priors into our framework.

\begin{table}[t]
    \centering
    \caption{\ourmodel is a general framework: In addition to NeRF, \ourmodel can benefit from data-driven priors of geometric information; here we use monocular depth estimation~\cite{yuan2022newcrfs} as an example. ``UN-Only:'' using the uncertainty network without augmentation of 3D structural information. Then we separately incorporate mono-depth or NeRF to it.}
    \label{tb:mono-depth}
    \vspace{-2mm}
    \resizebox{1\linewidth}{!}{
    \begin{tabular}{l|cccc}
    \toprule
    \multirow{2}{*}{Methods} & \multicolumn{4}{c}{mIoU} \\ & Divider & Ped Crossing & Boundary & All   \\ 
    \midrule
    Average Fusion       & 42.83  & 24.75 & 43.91 & 37.16 \\
    \midrule
    \ourmodel (UN-Only) & 47.38  & 31.11  & 49.53 & 42.67 \\
    \ourmodel (Mono-Depth)    & 48.04  & 32.96  & 50.08 & 43.69 \\
     \ourmodel (NeRF) & \textbf{48.15}  & \textbf{33.34}  & \textbf{50.28} & \textbf{43.92} \\
    \bottomrule
    \end{tabular}
    }
    \vspace{-3mm}
\end{table}


\subsection{Globally Consistent HD Map Generation}
Our offboard \ourmodel can handle numerous frames. Its application is to expand the range of HD map generation from a local region around the ego-vehicle to a global region covering all the input frames, which saves the labor in stitching multiple local predictions in the real world. Our global maps in Fig.~\ref{fig:global} demonstrate high fidelity for complex topology in two challenging scenes. While some regions do not match the ground truth, we argue that these regions fall outside the collected frames and perception ranges, which are beyond the scope of offboard algorithms.

\begin{table}[tb]
\caption{Performance of incorporating the {\em LiDAR modality} in \ourmodel, evaluated under the long-range setting on the validation set. As a {\em general} framework, \ourmodel can exploit multi-modality as input and improve the performance consistently and significantly.}\vspace{-2mm}
    \centering
    \resizebox{0.98\linewidth}{!}{
    \begin{tabular}{l|cccc}
    \toprule
    \multirow{2}{*}{Methods} & \multicolumn{4}{c}{mIoU (Long-range)} \\ & Divider & Ped Crossing & Boundary & All   \\ 
    \midrule
    Onboard & 41.63 & 27.13 & 41.65 & 36.80 \\
    MV-Map & \textbf{50.72} & \textbf{32.99} &  \textbf{54.63} &  \textbf{46.11} \\
    \bottomrule
    \end{tabular}
    }
    \vspace{-4mm}
    \label{tb:lidar}

\end{table}
\vspace{-1mm}
\subsection{Generalizability of \ourmodel with LiDAR}
\label{sec:lidar}

To demonstrate the generalizability of our framework, we analyze incorporating the LiDAR sensor into \ourmodel. In the main paper, we focused on cameras, because they contribute primarily to HD map generation as shown in HDMapNet~\cite{li2022hdmapnet}. On the other hand, LiDARs are also widely used for their accurate distance sensing and their ability to enhance localization, which motivates our investigation here. We first describe the design of utilizing the extra LiDAR modality and then analyze the results. Extra details are explained in Sec.~\ref{supsec:lidar}.
\vspace{-2mm}
\section{Conclusion}
\vspace{-2mm}
Regarding the infrastructure role of HD maps, we propose a novel \emph{offboard} HD map generation setup to address the unreliability of \emph{onboard} BEV perception. By removing the computation constraints, the models are allowed to reason all the frames altogether and construct multi-view consistent HD maps. Concretely, we propose an offboard HD map generation framework called \ourmodel. To address numerous frames, \ourmodel designs region-centric aggregation to unify the HD maps from all the frames. The key design is an uncertainty network that weighs the contribution of different frames and utilizes a Voxel-NeRF to provide multi-view consistent 3D structural information. Experiments validate that \ourmodel is scalable to large volumes of offboard data and significantly improves the HD map quality. We hope that our framework can become an effective augmentor for onboard algorithms and also inspire future research on offboard problems.

\mypar{Limitations and future work.} Although our Voxel-NeRF improves the offboard pipeline in a scalable way, several challenges still present, including moving objects in traffic scenes and exploiting data-driven priors for better geometric information. In addition, we seek to connect our work with auto-labeling and compare it with human annotation quality, so as to explore more potential applications such as autonomous vehicle navigation and urban planning.


\noindent{\textbf{Acknowledgement.} This work was supported in part by NSF Grant 2106825, NIFA Award 2020-67021-32799, the Jump ARCHES endowment, the NCSA Fellows program, the Illinois-Insper Partnership, and the Amazon Research Award. This work used NVIDIA GPUs at NCSA Delta through allocations CIS220014 and CIS230012 from the ACCESS program.}

\renewcommand\thesection{\Alph{section}}
\renewcommand\thetable{\Alph{table}}
\renewcommand\thefigure{\Alph{figure}}
\renewcommand\theequation{\Alph{equation}}

\setcounter{section}{0}
\setcounter{table}{0}
\setcounter{figure}{0}

{\small
\bibliographystyle{ieee_fullname}
\bibliography{egbib}
}

\clearpage

\setcounter{section}{0}
\setcounter{table}{0}
\setcounter{figure}{0}
\setcounter{equation}{0}
\renewcommand{\thesection}{A\arabic{section}}

\newcommand{\mainsec}[1]{\textcolor{blue}{#1}}
\def\ourmodel{MV-Map\xspace}
\def\ourmodelfull{Multi-view Map\xspace}

\renewcommand\thesection{\Alph{section}}
\renewcommand\thetable{\Alph{table}}
\renewcommand\thefigure{\Alph{figure}}
\renewcommand\theequation{\Alph{equation}}

\section*{Appendix}
Our supplementary contains the following contents:
\begin{enumerate}[leftmargin=*, noitemsep, nolistsep, label=(\Alph*)]
    \item \textbf{Demo video.} We provide a demo for offboard HD map generation in Sec.~\ref{supsec:video_demo}.
    \item \textbf{Voxel-NeRF details.} We explain the formulation of training and using Voxel-NeRFs in Sec.~\ref{supsec:voxel_nerf}.
    \item \textbf{Generalizability of \ourmodel with LiDAR.} In addition to the vision-oriented experimentation in the main paper, we show the generalizability of \ourmodel and incorporate it with the LiDAR modality in Sec.~\ref{supsec:lidar}.
    \item \textbf{Applications of Auto-labeling.} We validate the effectiveness of \ourmodel for auto-labeling, by using it to generate pseudo HD map labels in Sec.~\ref{supsec:auto_labeling}.
    \item \textbf{Additional quantitative results.} We supplement ablation studies, especially using additional onboard models, in Sec.~\ref{supsec:quantitative}.
    \item \textbf{Additional qualitative results.} The generated HD maps together with the reconstructed 3D structure via our Voxel-NeRF are visualized in Sec.~\ref{sec:vis-nerf}.
    \item \textbf{Implementation details.} We describe additional implementation details for reproducing our results in Sec.~\ref{supsec:implementation}. 
\end{enumerate}

\section{Demo Video}
\label{supsec:video_demo}

We provide a demo video at \url{https://youtu.be/SN14oTyMFrk} that showcases how our \ourmodel produces high-quality HD maps by fusing frames from diverse viewpoints. Notably, the video highlights the effectiveness of \ourmodel in {\em iteratively refining} complex road topologies and long road elements while dealing with frequent occlusions in urban traffic.

\section{Voxel-NeRF Details}
\label{supsec:voxel_nerf}
In this section, we introduce the details of optimizing our Voxel-NeRF and augmenting our \ourmodel with the encoded 3D structure (Sec.~\mainsec{4.3} of the main paper).

\subsection{NeRF optimization}
\label{supsec:nerf_optim}
We supervise our Voxel-NeRF in a way that is identical to standard NeRF models~\cite{nerfW, nerf, donerf, blocknerf}, by using a photometric loss between the rendered pixel color and the ground-truth color.

We first describe how NeRF infers the color of every pixel in this process. NeRF renders the color of an arbitrary pixel by accumulating the density and color information along the camera ray. Specifically, we denote the camera ray for the pixel as $\mathbf{r}$, which is unique for each pixel. By denoting the camera origin as $\mathbf{o}$ and the direction of $\mathbf{r}$ as $\mathbf{d}$, every 3D coordinate along the ray can be written as $\{\mathbf{o}+t\mathbf{d} | t \in\mathcal{R}^{+} \}$. The RGB color of the pixel comes from the integral along the ray $\mathbf{r}$:
\begin{align}
\label{supeq:nerf_rgb}
    \hat{\mathbf{C}}(\mathbf{r}) = \int_{t_{n}}^{t_{f}} T(t)\sigma(t)\mathbf{c}(t)\mathrm{d}t,    
\end{align}
where $t$ ranges from the near and far planes $t_n$ and $t_f$, $T(t) = \exp(-\int_{t_n}^{t}\sigma(\mathbf{o}+s\mathbf{d})\mathrm{d}s)$ models the accumulated transmittance along the ray from $t_{n}$ to $t$, and $\sigma$ and $\mathbf{c}$ denote the density and color encoded in NeRF, respectively. 

The photometric loss is a reconstruction loss between the RGB colors predicted by NeRF and from the ground-truth images:
\begin{align}
    \mathcal{L}_{\text{color}} = \mathbb{E}_{\mathbf{r}} \| \hat{\mathbf{C}}(\mathbf{r}) - \mathbf{C}(\mathbf{r}) \|^2_2,
\end{align}
where $\mathbf{C}(\mathbf{r})$ is the ground-truth RGB values extracted from the images. 

As discussed in Sec.~\mainsec{4.3} of the main paper, we further add a total-variance loss $\mathcal{L}_\text{TV}$ to guide the optimization of near-ground geometry. The final loss term is:
\begin{align}
    \mathcal{L} = \lambda_1\mathcal{L}_{\text{color}} + \lambda_2\mathcal{L}_{\text{TV}},
\end{align}
where $\lambda_1$ and $\lambda_2$ are trade-off hyper-parameters.

\subsection{NeRF ray casting}
\label{supsec:ray_casting}
In Sec.~\mainsec{4.3} of the main paper, we show how we incorporate the multi-view geometry in Voxle-NeRF with our uncertainty network. The {\em key operator} is to reconstruct the position of the nearest surface for each voxel by ray-casting through the corresponding image pixel. We achieve this by rendering the {\em termination depth} through volume rendering.

Specifically, for every camera ray represented in the form of $\{\mathbf{o}+t\mathbf{d} | t \in \mathcal{R}^{+}\}$ (explained in Sec.~\ref{supsec:nerf_optim}), the termination depth of the ray $\hat{D}(\mathbf{r})$ is:
\begin{align}
\label{supeq:depth}
    \hat{D}(\mathbf{r}) = \int_{t_{n}}^{t_{f}} T(t)\sigma(t)\mathrm{d}t.
\end{align}
Similar to Eqn.~\ref{supeq:nerf_rgb}, $t$ ranges from the near and far planes $t_n$ and $t_f$, $T(t) = \exp(-\int_{t_n}^{t}\sigma(\mathbf{o}+s\mathbf{d})\mathrm{d}s)$ models the accumulated transmittance along the ray from $t_{n}$ to $t$, and $\sigma$ denotes the density in NeRF. 

\section{Generalizability of \ourmodel with LiDAR}
\label{supsec:lidar}

In this section, we provide the model design details of incorporating \ourmodel with LiDAR modality as described in Sec.~\mainsec{5.6} (main paper).

\mypar{Onboard model.} We modify the original image-based onboard model by adding a branch of LiDAR encoder with PointPillar~\cite{lang2019pointpillars} to generate the BEV feature maps from the point clouds. The BEV feature maps generated by the LiDAR encoder are later stacked with the image-based BEV features to form the final BEV features.

\mypar{Uncertainty network.} The architecture of the uncertainty network remains unchanged when integrating the LiDAR sensor, as our offboard fusion pipeline is \emph{agnostic} to the upstream BEV perception modules.

\mypar{Voxel-NeRF.} In addition to optimizing the Voxel-NeRF with the photometric loss and total-variance loss as in Sec.~\mainsec{4.3} (main paper), we further leverage the point clouds to improve NeRF. Specifically, we follow DS-NeRF~\cite{deng2021depth} and apply an extra depth loss term:
\begin{align}
    \mathcal{L}_{\text{depth}} = \mathbb{E}_{\mathbf{r}} \| \hat{D}(\mathbf{r}) - D(\mathbf{r}) \|_2^2,
\end{align}
where $\hat{D}(\mathbf{r})$ is the rendered termination depth in Eqn.~\ref{supeq:depth}, and ${D}(\mathbf{r})$ is the depth of LiDAR points. Note that point clouds are sparser than image pixels, so we project LiDAR points onto the images and only apply the above loss term to the pixels that correspond to LiDAR points for supervision.

Our final training loss for Voxel-NeRF combines the photometric, total-variance, and depth losses when the LiDAR modality is available:
\begin{align}
\setlength{\abovecaptionskip}{2pt}
\setlength{\belowcaptionskip}{2pt}
    \mathcal{L} = \lambda_1\mathcal{L}_{\text{color}} + \lambda_2\mathcal{L}_{\text{TV}} + \lambda_3 \mathcal{L}_{\text{depth}},
\end{align}
where $\lambda_1$, $\lambda_2$, and $\lambda_3$ are trade-off hyper-parameters.

\mypar{Results.} Table~\mainsec{5} (main paper) summarizes the result of \ourmodel with LiDAR, which again significantly outperforms the onboard model. In addition, due to leveraging the additional modality, \ourmodel with both camera and LiDAR achieves larger improvement, compared with the unimodal model result with only camera shown in Table~\mainsec{1} This result serves as further evidence that our framework is capable of adapting to multi-modality and achieving improved performance.
\begin{table}[tb]
\caption{Comparison between onboard models trained with either ground-truth labels (GT) or pseudo-labels generated by our \ourmodel (PL). The model trained with our pseudo-labels achieves comparable performance. This validates the high quality of HD maps generated by \ourmodel and further supports its effectiveness for auto-labeling.}\vspace{-2mm}
    \centering
    \resizebox{0.95\linewidth}{!}{
    \begin{tabular}{l|cccc}
    \toprule
    \multirow{2}{*}{Label} & \multicolumn{4}{c}{mIoU (Validation set, Long-range)} \\ & Divider & Ped Crossing & Boundary & All   \\ 
    \midrule
    PL (Ours)  & \textbf{38.99}  & 25.15  & \textbf{38.68} & \textbf{34.27} \\
    GT & 38.89  & \textbf{25.40}  & 38.16 & 34.15 \\
    \bottomrule
    \end{tabular}
    }
    \vspace{-2mm}
    \label{tb:auto-label}

\end{table}

\vspace{-2mm}
\section{Applications of Auto-labeling}

\label{supsec:auto_labeling}
The experimental results in the main paper show that \ourmodel generates high-quality HD map labels. This indicates that our method is an effective {\em auto-labeling} strategy, which can potentially serve as a substitute for human labeling and thus support downstream applications. To further assess the quality of these labels, which we refer to as ``\emph{pseudo-labels},'' we conduct an experiment by training a new onboard model with pseudo-labels and comparing its efficacy with that trained with ground-truth labels.

To this end, we follow the \emph{semi-supervised learning} experimental setup introduced in offboard 3D detection~\cite{qi2021offboard}. We use 50 out of 700 sequences on the training set of nuScenes~\cite{caesar2020nuscenes} to train our uncertainty network and deploy it to infer the HD map labels for the remaining 650 sequences on the training set. We then train an onboard model {\em from scratch} on these 650 sequences with either the ground-truth labels or pseudo-labels from \ourmodel. 

The result in Table~\ref{tb:auto-label} shows that the model trained with pseudo-labels achieves comparable performance to that trained with ground-truth labels. This suggests that our auto-labeling approach is effective for supporting semi-supervised training. It is worth noting that our pseudo-labeling performs slightly better, likely because it helps to reduce over-fitting as evidenced by that on the {\em training} set using ground-truth labels results in 2.5\% higher mIoU over pseudo-labels. Based on the high quality of the generated HD maps, our auto-labeling pipeline has the potential to be useful for other BEV perception tasks that involve traffic elements, such as BEV segmentation and lane detection. We leave such investigation as interesting future work.

\section{Additional Quantitative Results}
\label{supsec:quantitative}
\subsection{Impact of Training-time Frame Number}

As described in Sec.~\mainsec{4.4}, we train the uncertainty network on clips with a fixed number of frames due to GPU capacities but later apply it to \emph{sequences with unbounded lengths}. We analyze how the training-time frame number impacts the performance of the uncertainty network. In Table~\ref{tb:fusion-frames}, we demonstrate that {\em increasing the number of frames is beneficial to the fusion performance}, as the uncertainty network has access to more diverse viewpoints for fusion during the training time. Moreover, increasing from 3 to 5 frames has a significant gain in performance, while increasing from 5 to 7 frames only has marginal improvement. Our experiments in the main paper use 5-frames for training to balance the performance and computation cost. 

\begin{table}[tb]
\caption{Performance of \ourmodel with varied training-time frame numbers. Given that the performance saturates at 5 frames, we adopt 5 frames for training to best trade-off accuracy and computational cost. Note that during inference, we apply \ourmodel to sequences with unbounded lengths.}\vspace{-2mm}
    \centering
    \resizebox{0.99\linewidth}{!}{
    \begin{tabular}{c|cccc}
    \toprule
    \multirow{2}{*}{\#Frames} & \multicolumn{4}{c}{mIoU (Long-range)} \\ & Divider & Ped Crossing & Boundary & All   \\ 
    \midrule
    3        & 47.64  & 32.36  & 49.67 & 43.22 \\
    5        & 48.15  & 33.34  & \textbf{50.28} & 43.92 \\
    7        & \textbf{48.23}  & \textbf{34.31}  & 50.11 & \textbf{44.22} \\
    \bottomrule
    \end{tabular}
    }
    \vspace{-2mm}
    \label{tb:fusion-frames}

\end{table}

\subsection{Generalizing to Additional Onboard Models}
We demonstrate that the region-centric fusion approach in our \ourmodel is {\em generalizable to other onboard BEV perception models}. In addition to the SimpleBEV~\cite{harley2022simple} and VectorMapNet discussed in Sec.~\textcolor{blue}{4} of the main paper, we adopt HDMapNet~\cite{li2022hdmapnet}. Following our experiment in Sec.~\ref{supsec:lidar}, we incorporate the HDMapNet encoder with the LiDAR point clouds. As shown in Table~\ref{tb:supp-encoder}, \ourmodel significantly improves upon both the HDMapNet~\cite{li2022hdmapnet} and VectorMapNet~\cite{liu2022vectormapnet} result, thus supporting the generalizability of \ourmodel.

\begin{table}[tb]
\caption{MV-Map generalizes to other onboard models. Using HDMapNet~\cite{li2022hdmapnet} as our onboard model, \ourmodel is consistently effective and significantly improves the HD map quality.} \vspace{-2mm}
    \centering
    \resizebox{0.99\linewidth}{!}{
    \begin{tabular}{c|cccc}
    \toprule
    \multirow{2}{*}{Methods} & 
    \multicolumn{4}{c}{mIoU} \\ & Divider & Ped Crossing & Boundary & All   \\ 
    \midrule
    HDMapNet & 46.20  & 24.38  & 56.99 & 42.52 \\
    +\ourmodel & \textbf{49.82}  & \textbf{29.83}  & \textbf{58.54} & \textbf{46.06} \\
    \rowcolor{lightgreen}
    $\triangle$ mIoU&\textcolor{gray}{+3.62}&\textcolor{gray}{+5.45}&\textcolor{gray}{+1.55}&\textcolor{gray}{+3.54}\\
    \bottomrule
    \end{tabular}
}
    \vspace{-2mm}
    \label{tb:supp-encoder}
\end{table}

\subsection{Sensitivity to KL divergence loss}
To further investigate the sensitivity of \ourmodel's performance to the weight of the KL divergence loss (Sec.~\mainsec{4.2}, main paper), we conduct ablation experiments with different settings of $\omega$. Compared to the original setting of $\omega$ = 0.1, setting $\omega$ = 0.2 results in an increase of 0.25 in mIoU, while setting $\omega$ = 0.05 causes a small decrease of 0.04 in mIoU. These minor variations in mIoU with sizable changes in $\omega$ indicate that \ourmodel's performance is fairly robust to the exact weight of the KL divergence loss.

\begin{table}[t]
    \centering
    \caption{Comparison between fusing BEV feature maps $F_i$ and semantic maps. We choose to fuse semantic maps in Sec.~\mainsec{4.2} (main paper) because of its better performance. }
    \label{tb:feature-fusion}
    \resizebox{0.95\linewidth}{!}{
    \begin{tabular}{l|cccc}
    \toprule
    \multirow{2}{*}{Fusion} & \multicolumn{4}{c}{mIoU} \\ & Divider & Ped Crossing & Boundary & All   \\ 
    \midrule
    Onboard        & 39.30  & 26.44  & 39.10 & 34.95 \\
    \midrule
    BEV feature    & 45.88  & 33.06 & 46.38 & 41.77 \\
    Semantic       & \textbf{48.15}  & \textbf{33.34}  & \textbf{50.28} & \textbf{43.92} \\
    \bottomrule
    \end{tabular}
    }
    \vspace{-4mm}
\end{table}

\subsection{Fusing semantics versus BEV features.} Our region-centric framework performs weighted averages over the semantic maps $S_i$ instead of the BEV features $F_i$. In Table~\ref{tb:feature-fusion}, we justify our design choices, where fusing BEV features is worse than fusing semantic maps. The main reason is the domain shift between training and inference when we have numerous input frames of offboard data. Furthermore, fusing BEV features is also less practical which requires significantly more disk space to store high-dimensional features. 
\section{Additional Qualitative Results}
\label{sec:vis-nerf}

\begin{figure}[tb]
    \centering
    \includegraphics[width=0.95\linewidth]{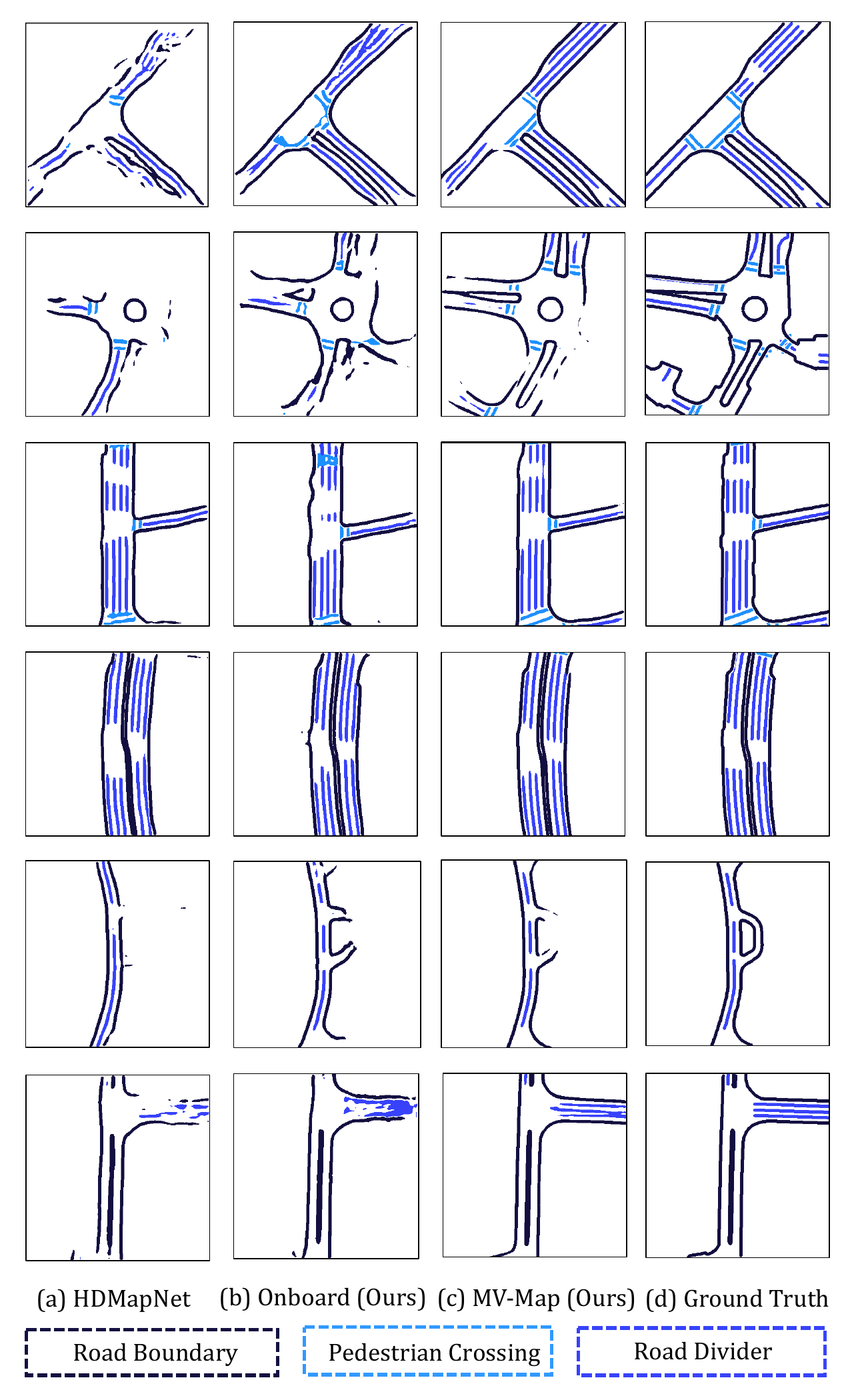}
    \vspace{-2mm}
    \caption{Qualitative results for HD map generation. We compare the generated results from HDMapNet~\cite{li2022hdmapnet}, our onboard model, and the offboard fused results from \ourmodel. Compared with other approaches, our MV-Map achieves better fidelity for complex road topology.}
    \vspace{-4mm}
    \label{fig:sup_global}
\end{figure}

\subsection{HD-Map Visualization}

We provide more qualitative results on HD map generation in Fig.~\ref{fig:sup_global}, in addition to our visualizations in Fig.~\mainsec{5} and Fig.~\mainsec{8} of the main paper. Compared with onboard approaches, our offboard \ourmodel significantly improves the quality of HD maps for complex structures.

\subsection{Voxel-NeRF Visualization}
\begin{figure}[t]
    \centering
    \includegraphics[width=0.99\linewidth]{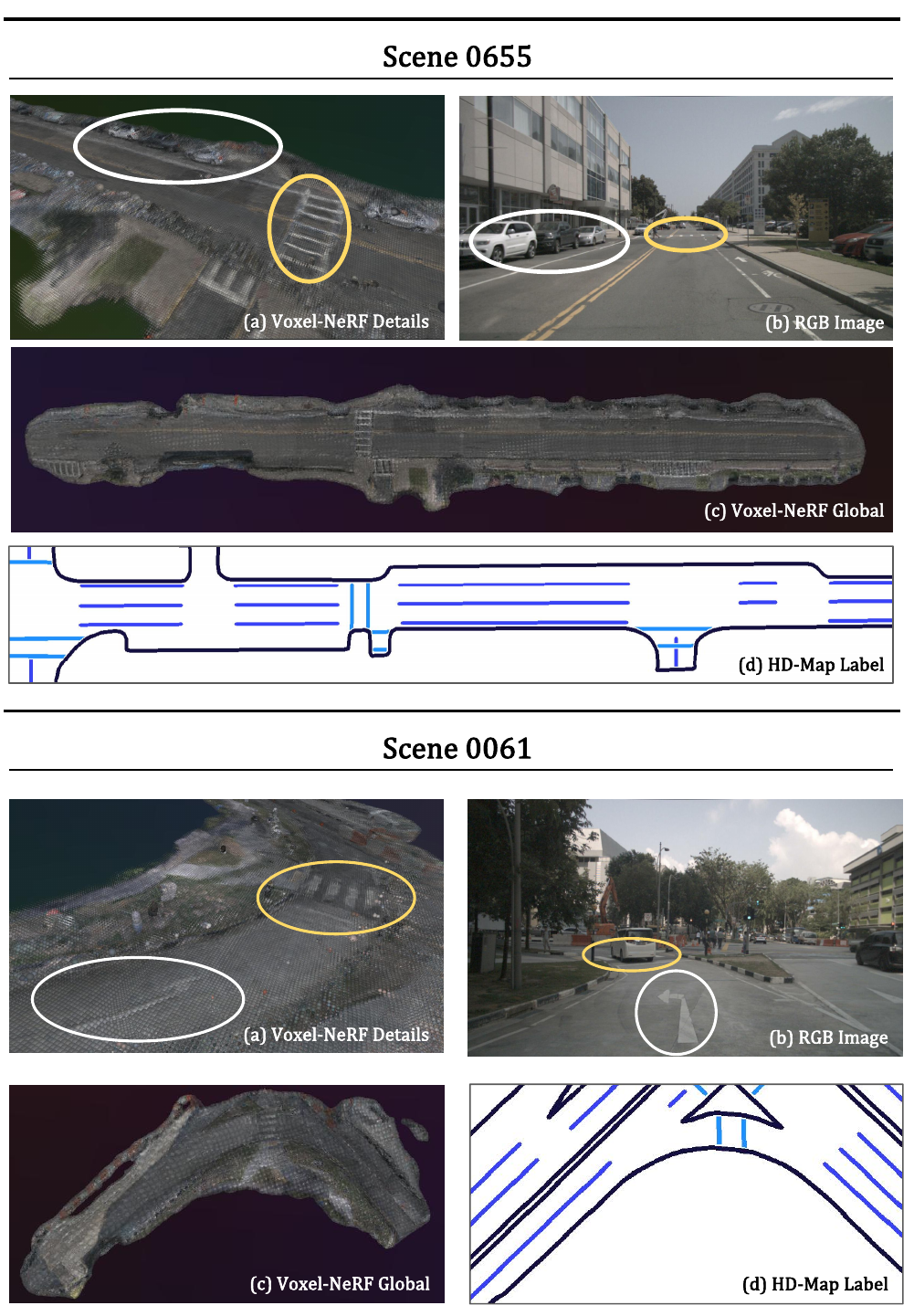}
    \vspace{-2mm}
    \caption{Visualization of reconstruction results by Voxel-NeRF on the scene 0655 and 0061 from nuScenes. \textbf{(a)} NeRF's results in the highlighted regions; \textbf{(b)} images captured by the ego vehicle; \textbf{(c)} NeRF's results for the whole scene; \textbf{(d)} ground-truth HD map labels. As highlighted here, our Voxel-NeRF optimizes the 3D structure of the whole scene with high quality and multi-view consistency.}
    \vspace{-3mm}
    \label{fig:nerf}
\end{figure}

We provide more visualization results of our Voxel-NeRF to indicate its capability of encoding multi-view consistency. In Fig.~\ref{fig:nerf}, we show the reconstructed 3D structure of the scenes by our Voxel-NeRF model, by converting the diffuse color and opacity of every voxel to a colored point cloud. Qualitative results demonstrate that our Voxel-NeRF successfully optimizes a high-resolution scene representation with multi-view consistency. 

\section{Implementation Details}
\label{supsec:implementation}
We provide the detailed hyper-parameters and procedures to reproduce \ourmodel as described in Sec.~\mainsec{4.4} and Sec.~\mainsec{5.1} (main paper), as well as Sec.~\ref{supsec:lidar}.

\subsection{Onboard Model} 
\label{supsec:onboard}

As SimpleBEV~\cite{harley2022simple} was not originally proposed for HD map construction (despite its strong performance), we inherit their encoder design and train our own onboard models. 

\mypar{Backbone.} Same as~\cite{harley2022simple}, we adopt ResNet-50~\cite{he2016deep} as the backbone to extract the feature maps for 6 surrounding images per frame on nuScenes. The third and final stages of the ResNet output are used. We apply an additional convolution layer to generate the final feature map with a length and width of 1/8 compared to the original size of the images.

\mypar{Feature lifting.} For feature lifting, we use a sampling-based BEV encoder to lift the 2D image feature into BEV space. We first construct a local 3D voxel grid shaped $400\times 400\times 6$ around the ego vehicle. The voxel grid size is 0.15m for the short-range (60m $\times$ 30m) setting and 0.25m for the long-range (100m $\times$ 100m) setting. Then, for each grid point, we acquire its positions on the image plane with intrinsic and extrinsic matrices, bi-linear sample the image features, and fill them back into each voxel grid. Finally, we reduce the voxel features into a BEV feature map with an additional voxel encoder to make it a 2D BEV feature. During this process, the height range of the sampled voxel grid in our BEV encoder is -4m to 2m relative to the sensor origin. The final output BEV feature map shapes 128 $\times$ 400 $\times$ 400, where 128 is the feature dimension, and 400 is the length and width of the BEV feature map.

\mypar{Decoder.}
To maintain generality and comparability, we used the same decoder as HDMapNet~\cite{li2022hdmapnet}. The main structure contains three blocks from ResNet18~\cite{he2016deep} to generate the final prediction.

\mypar{Loss function.} 
We use Focal loss~\cite{lin2017focal_loss}, which is a dynamically scaled cross-entropy loss as our segmentation loss to solve the imbalance distribution between the 
 most common road label and the crossing label that is relatively scarce and hard to learn:
\begin{align}
     \textbf{FL}(p_t) = -\alpha_t(1-p_t)^{\gamma}\log(p_t),
\end{align}
where we set the factor $\alpha_t = 1$ and $\gamma = 2$. During training, the scaling factor can reduce the impact of dominant categories (\emph{e.g.}, road segments) and increase the loss assigned to challenging ones (\emph{e.g.}, pedestrian crossing).
 
\mypar{Training.} During training, we initialize the backbone ResNet-50 from the ImageNet1k~\cite{deng2009imagenet} pretrained checkpoint. It is then trained for 16 epochs with an AdamW~\cite{loshchilov2017decoupled} optimizer, with an initial learning rate of 1e-3 under a 1-cycle schedule and focal loss~\cite{lin2017focal_loss} as loss function. We train our model with 4$\times$A100 GPUs with 2 samples per GPU. The total training process takes around 10 hours.

\subsection{Voxel-NeRF}

\mypar{Architecture.}
Our Voxel-NeRF models are trained with a fixed voxel size of 0.5m. The height range of our voxel is set to -4m to 2m relative to the height value of sensor origins. The near plane and the far plane of our NeRF model are 0.1m and 64m, respectively. The RGB network has a width of 128 and a depth of 3 layers. Within our model, each voxel encodes the feature with dimension 12. The first 3 channels represent the diffuse color, and the rest 9 channels concatenate the viewing directions to decode the final RGB color $c$ with an RGB MLP.

\mypar{Training.} We reconstruct all 850 scenes in nuScenes and train 30,000 iterations with AdamW~\cite{loshchilov2019adamw} optimizer and 1e-3 learning rate for each scene. Please note that our training is \emph{without} the coarse-to-fine strategy proposed in~\cite{dvgo}. As our scene scale is predefined and does not need the coarse stage to find a tight bounding box for further optimization. When we use the total-variance loss (Sec.~\mainsec{4.3}, main paper), the balance loss weights $\lambda_1$ is 1 and $\lambda_2$ is 1e-5. The training process takes around 15 minutes for each scene on a single A40 GPU.

\subsection{Uncertainty Network}
\label{supsec:uncertainty}

\mypar{Architecture.} We show our uncertainty network design in Fig.~\mainsec{4} (main paper). It has a U-Net-like architecture which takes in the 128-channel BEV feature map with channels, and a 6-channel augmented input from NeRF (as Sec.~\mainsec{4.3}, main paper), and outputs a final 128-channel feature map. Then the per-pixel confidence weight and the KL divergence prediction are output by two independent 1$\times$1 convolution layers with kernel size 1.

\mypar{Training.} Due to our storage constraints, we train the uncertainty network on a small subset (50 scenes) of the full training set for 5 epochs, but we manage to evaluate the full validation set of nuScenes with 150 sequences, which enables a fair comparison with other methods. As explained in Sec.~\mainsec{4.4} and Sec.~\mainsec{5.1} (main paper), we train the uncertainty network on short video clips with 5 frames and deploy it to all the frames in a nuScenes sequence during the inference time. A key detail to handle the varying sequence lengths for inference time is to set the \emph{batch size} to 5 during the training time. The final loss is a weighted sum of the segmentation loss and the auxiliary KL divergence loss (Sec.~\mainsec{4.2}, main paper), the loss weights are 1 and 0.1, respectively.
The network is trained with an AdamW \cite{loshchilov2017decoupled} optimizer with a learning rate of 1e-3. The training process takes around 30 minutes on a single A100 GPU.

\subsection{\ourmodel with LiDAR}
\label{supsec:lidar_details}

For the implementation of the LiDAR \ourmodel, we maintain the hyperparameters of the onboard model and the uncertainty network at the same values as the unimodal model. The onboard model's training process requires around 12 hours to complete using 4$\times$A100 GPUs, while the uncertainty network training process takes an additional hour using a single A100 GPU.

As discussed in Sec.~\ref{supsec:lidar}, when training with LiDAR signals, our Voxel-NeRF applies an extra depth loss term with $\lambda_3 = 0.1$. We keep other hyperparameters the same and train our Voxel-NeRF for 30,000 iterations for each scene, which takes 15 minutes on a single A100 GPU.

\subsection{\ourmodel with Monocular Depth}
We experiment using monocular depth for uncertainty network in Sec.~\mainsec{5.3} (main paper). Specifically, we replace the termination depth from NeRF (Eqn.~\ref{supeq:depth}) with the depth generated by {\em NeWCRFs}~\cite{yuan2022newcrfs} to estimate the 3D positions of surface points. The other implementation details are untouched. 

\clearpage

\end{document}